\documentclass[10pt,twocolumn,letterpaper]{article}

\usepackage[pagenumbers]{wacv}%

\usepackage{enumitem}
\usepackage{CJKutf8}%

\usepackage{tikz}
\usepackage{soul}

\usetikzlibrary{calc,positioning,arrows.meta}

\definecolor{wacvblue}{rgb}{0.21,0.49,0.74}
\usepackage[pagebackref,breaklinks,colorlinks,allcolors=wacvblue]{hyperref}
\usepackage[capitalize]{cleveref}

\def\wacvPaperID{3699}%
\def\confName{WACV}
\def\confYear{2027}

\newcommand{\Tone}{$\tau_1$\xspace}
\newcommand{\Ttwo}{$\tau_2$\xspace}
\newcommand{\Tthree}{$\tau_3$\xspace}
\newcommand{\rcond}{\rho_{\mathrm{cond}}}
\newcommand{\rdiv}{\rho_{\mathrm{div}}}
\newcommand{\rfaith}{\rho_{\mathrm{faith}}}
\newcommand{\rfaithref}{\rho_{\mathrm{faith,ref}}}

\title{Conditional Collapse in Sign Language Production:\\
       A Diagnostic and a Coverage Argument}

\author{Rui Hong \qquad Jana Ko\v{s}eck\'a\\
George Mason University\\
{\tt\small \{rhong5, kosecka\}@gmu.edu}}

\begin{document}
\maketitle
\begin{abstract}
The goal of Sign Language Production (SLP) is to translate natural language text into corresponding sign language movements performed by a virtual avatar. The quality of the
generated motion is typically evaluated by a motion-space Fr\'echet
distance (FID) and back-translation BLEU score on benchmarks
such as How2Sign~\cite{duarte2021how2sign}. The existing approaches can improve both metrics substantially while the
underlying generator fails to represent the input sentence 
--- a failure we call \emph{conditional collapse}. In this work we
propose to evaluate the generated motion at three independent
levels: (\Tone) initial-pose conditioning, (\Ttwo) output diversity,
and (\Tthree) target faithfulness. We compute these as
pairwise-distance ratios using latent representations of a frozen
motion autoencoder (MoAE), judged against each dataset's own
GT-vs-GT reference.
We evaluate diffusion and regression SLP models on How2Sign, including
a re-implemented Neural Sign Actors, and show that \Tthree{}
faithfulness is never attained, while FID varies by nearly two
orders of magnitude and is uncorrelated with faithfulness. We then
apply the same models and diagnostic to Phoenix-2014T, CSL-Daily,
and gloss-level ASL3DWord and show that faithfulness varies 
proportionally to how densely the training sentences cover the test conditions
(\emph{conditioning coverage}) and not corpus size. The smaller but
densely covered Phoenix-2014T and ASL3DWord datasets generate faithful 
sign motion, while CSL-Daily follows the input text yet falls short of
sign language faithfulness. Using the largest, open-domain How2Sign 
dataset, the models collapse at every scale. We introduce two measurements of the text data 
that rank the corpora in the order of how densely the training
sentences fill the embedding space, and how close each test
sentence lies to its nearest training sentence.
\end{abstract}

\section{Introduction}
\label{sec:intro}

Sign Language Production (SLP) systems generate avatar sign language
motion from
text~\cite{saunders2020progressive,baltatzis2023nsa,xie2024g2pddm,yin2024t2sgpt,zuo2025soke}
and report steady benchmark gains. Reliable
evaluation requires metrics that respond to whether each output is
faithful to its \emph{specific} input, not merely plausible under
the training distribution. We study this requirement in
sentence-level SLP, where systems on
How2Sign~\cite{duarte2021how2sign} report improving Fr\'echet
distance (FID) over motion features and back-translation (BT) BLEU.
We argue that both headline metrics can rise while the generator
remains essentially text-independent. FID rewards an
anatomically valid mean motion that ignores the input over a model
that follows the input but produces slightly noisier motion; it can
thus prefer a collapsed generator over a faithful one, as we observe
on How2Sign (\cref{sec:fid_disconnect}). On How2Sign, BT-BLEU is bounded by the BT
model. Without pretraining, How2Sign alone does not train a usable
translator, and the BT model collapses to high-frequency boilerplate,
reflecting token-frequency overlap rather than motion--text
correspondence.

Beyond \emph{detecting} this failure, we ask \emph{what drives it}.
Our answer is a coverage argument: whether a generator learns to
condition on text is governed by \emph{conditioning coverage} ---
whether semantically close training sentences recur often enough in
the conditioning space for the text-to-motion correspondence to be
learnable. We show a corpus with $31$K paired sentences
collapsing while a $7$K-sentence corpus with denser coverage does
not.

\textbf{Contributions.}
{\bf (i)} Given any text-to-motion model and a held-out test set, we
encode both ground-truth motions and model generations through a
frozen motion autoencoder MoAE (a small transformer) trained once on
the target dataset's ground-truth motions, compressing each motion
into a compact latent vector and reconstructing it. We then compute
three pairwise-distance ratios in that space, varying the sentence
conditioning and comparing the generated and ground-truth motions;
faithfulness is judged against the dataset's own measured GT-vs-GT
reference rather than a fixed threshold (\cref{sec:reference}), and
a frame-level DTW check against a human--human alignability baseline
sharpens the verdict (\cref{sec:framelevel}).
{\bf (ii)} We identify \emph{three indicators} of conditional collapse
at different levels: an initial-pose level (\Tone), an output
diversity level (\Ttwo), and a faithfulness level (\Tthree) that can
be attained \emph{independently}, with visual and quantitative
evidence (\cref{sec:three_layer}).
{\bf (iii)} We run the \emph{same}
architectures, recipe, and diagnostic across four datasets spanning a
coverage spectrum: open-domain How2Sign, weather-domain
Phoenix-2014T~\cite{forster2014phoenix}, daily-domain
CSL-Daily~\cite{zhou2021csldaily}, and isolated-gloss
ASL3DWord~\cite{dong2024signavatar}, supported by measurements of the text data alone (\cref{sec:coverage}). Controls (text shuffling, frozen-encoder
swaps) rule out the text encoder; a retrained Progressive
Transformers~\cite{saunders2020progressive} collapsing on
densely-covered Phoenix shows coverage is \emph{necessary but not
sufficient}.

The failures we document are diagnosable: report the diagnostic
ratios alongside FID and BT-BLEU, judge them against per-dataset
GT-vs-GT references, and be transparent about the coverage
properties of the training corpus.

\section{Related Work}
\label{sec:related}

\noindent\textbf{Sign Language Production.}
Existing sentence-level SLP approaches fall into two families. \emph{Regression based models}, such as Progressive Transformers (PT)~\cite{saunders2020progressive} decode
poses autoregressively with a learned step counter;
\citet{huang2021fast} decode non-autoregressively with a
graph-convolutional pose generator; T2S-GPT~\cite{yin2024t2sgpt}
generates dynamic-length VQ pose codes autoregressively.
A representative \emph{Diffusion-based model}, MDM~\cite{tevet2023mdm}, is a transformer denoiser
on frozen CLIP conditioning; G2P-DDM~\cite{xie2024g2pddm} runs
discrete diffusion in a VQ latent; Neural Sign Actors
(NSA)~\cite{baltatzis2023nsa} is a GNN$+$LSTM denoiser on SMPL-X
parameters. \citet{hong2026signphonology} report usable MDM-style
generation at the \emph{gloss level} on
ASL3DWord~\cite{dong2024signavatar} (a WLASL~\cite{li2020word}
subset); 
Signs as Tokens
(SOKE)~\cite{zuo2025soke} generates part-wise discrete sign tokens
with a pretrained language model across How2Sign, Phoenix-2014T,
and CSL-Daily, augmented by \emph{retrieving word-level signs from
external dictionaries} --- which we read as independent evidence for
the coverage account of \cref{sec:coverage}: word-level signs are
injected because sentence-level pairs alone do not cover the
conditions. We use SOKE's released SMPL-X fittings as paired data
(its generator weights are unreleased).

\noindent\textbf{Sign Language Translation and BT evaluation.}
SLP approaches are commonly scored by back-translating generated motion
and reporting BLEU translation score. On How2Sign, no-pretraining translators plateau
near BLEU-4 $1.2$~\cite{youtubeasl2023}, reaching $8.03$ with I3D
features~\cite{tarres2023h2s} and $\sim$$12$ only after $11$K-hour
pretraining~\cite{youtubeasl2023}; our own BT model plateaus at
$\approx 2.5$ even on GT motion. PT~\cite{saunders2020progressive}
and NSA~\cite{baltatzis2023nsa} report scores of $10.51$ and $13.12$ without
pretraining, while independent retrainings place PT at
$1.6$--$4.4$~\cite{huang2021fast,xie2024g2pddm,yin2024t2sgpt} and
note its dependence on GT timing.

\noindent\textbf{Evaluation critiques.}
FID's limitations for conditional generation are known in
text-to-image~\cite{kynkaanniemi2022role} and
text-to-motion~\cite{guo2022humanml3d} domains, motivating paired R-precision
and multimodality metrics. Our diagnostic targets the
\emph{conditional-mean collapse} blindness dominant in SLP and
calibrates its verdicts against per-dataset GT-vs-GT references.
Pose-space DTW distances (DTW-MJE) are standard on
Phoenix-2014T~\cite{saunders2020progressive,huang2021fast,yin2024t2sgpt}
but are reported without a human--human baseline, which
\cref{sec:framelevel} supplies.
\citet{jiang2025meaningful} meta-evaluate pose-based sign language
metrics by correlation with human judgment; we instead diagnose
\emph{why} generators fail.

\section{Diagnostic Approach}
\label{sec:diagnostic}

Let $f_\theta$ be any Sign Language Production (SLP) model, regression
or diffusion-based, that takes a sentence in natural language and
generates a sequence of human poses. Consider
$\{(t_i, m_i)\}_{i=1}^{N}$ the full test set, where $t_i$ is the
$i$-th text input and $m_i \in \mathbb{R}^{T \times D}$ its
ground-truth motion clip.
Let $\phi: \mathbb{R}^{T \times D} \to \mathbb{R}^{d}$ be a frozen
motion autoencoder MoAE, with latent dimension $d = 256$.
We use $T = 100$ frames, per-frame
$D = 44 \text{ upper-body joints} \times 3 = 132$ pelvis-relative
$3$D coordinates, and train one MoAE per dataset on its train
motions only (see architecture in Appendix~D). Define
$z_i := \phi(f_\theta(t_i; s_1))$ as the latent representation of SLP model
output, and $z_i^{*} := \phi(m_i)$, the MoAE encoded GT motion for the
associated text sentence. We compute the following pairwise statistics:
\begin{align}
d_{\mathrm{inter}} &= \mathbb{E}_{i \neq j} \,
   \| z_i - z_j \|_2,                          \\
d_{\mathrm{intra}} &= \mathbb{E}_i \,
   \big\| \phi(f_\theta(t_i; s_1)) - \phi(f_\theta(t_i; s_2)) \big\|_2.
\end{align}
Here $s_1 \ne s_2$
are two independent sampling seeds of a stochastic (diffusion)
generator; a deterministic (regression) model takes no seed, so
$f_\theta(t_i; s_1) = f_\theta(t_i; s_2)$.
\begin{align}
d_{\mathrm{inter\text{-}GT}} &= \mathbb{E}_{i \neq j} \,
   \| z_i^{*} - z_j^{*} \|_2,                  \\
d_{\mathrm{pair}} &= \mathbb{E}_i \,
   \| z_i - z_i^{*} \|_2,                       \\
d_{\mathrm{pair\text{-}rand}} &= \mathbb{E}_i \,
   \| z_i - z_{\pi(i)}^{*} \|_2.
\end{align}
Here $\pi$ randomly shuffles $\{1,\dots,N\}$, leaving no index in
place, so $z_{\pi(i)}^{*}$ is the GT motion of a different test
sample.

Only $d_{\mathrm{inter}}$ and $d_{\mathrm{pair}}$ capture how the
output pose is affected by the input sentence and whether it lands
on the correct target;  $d_{\mathrm{intra}}$ isolates pure
sampling noise, so input-driven variation can be told apart from it;
$d_{\mathrm{inter\text{-}GT}}$ measures how much the GT motions of
different sentences naturally differ, so a small
$d_{\mathrm{inter}}$ signals collapse only when it is small compared
to this natural spread; and $d_{\mathrm{pair\text{-}rand}}$ measures
whether the generated SLP output is closer to another example
other than its own GT.

\textbf{Three diagnostic ratios.} From these statistics we form three unitless
quantities: a \emph{condition} ratio (input-driven variance vs.\
sampling noise), a \emph{diversity} ratio (output variance vs.\ GT
cross-sentence variance), and a \emph{faithfulness} ratio (target
alignment vs.\ a random-GT baseline):
\begin{align}
\rcond  &= d_{\mathrm{inter}} / d_{\mathrm{intra}}, \\
\rdiv   &= d_{\mathrm{inter}} / d_{\mathrm{inter\text{-}GT}}, \\
\rfaith &= d_{\mathrm{pair}} / d_{\mathrm{pair\text{-}rand}}.
\end{align}
Since $\rcond$ requires a stochastic sampling, a deterministic model has
$d_{\mathrm{intra}} = 0$ (no seed dependence), making $\rcond$
undefined. The other two ratios only need a single generation per
sentence, so they still apply; for deterministic models we report
$\rdiv$ and $\rfaith$ only.

FID measures a Fr\'echet distance
between two unconditional Gaussians fit to feature distributions: a
model whose outputs cluster around the GT marginal mean attains small
FID even when individual outputs $f_\theta(t_i)$ are not faithful
motion representing their input $t_i$.
The guideline of $\rfaith$ approaching $1.0$ captures
this failure mode directly. The diagnostic requires no new training
(it reuses one frozen MoAE and $O(N^2)$ pairwise distances on
the full test split) and costs at most one extra forward pass per
sentence (the second seed for $d_{\mathrm{intra}}$; none for
deterministic models).

\subsection{Per-dataset GT-vs-GT reference}
\label{sec:reference}

\begin{table}[t]
\centering
\scriptsize
\setlength{\tabcolsep}{2pt}
\caption{GT-vs-GT reference per dataset: same-text groups,
renditions, same-/different-text pair counts, and $\rfaithref$
(\cref{eq:rfaith_ref}); protocol and latent distances in
Appendix~E.}
\label{tab:reference}
\begin{tabular}{lrrrrc}
\toprule
Dataset & groups & rend. & same & diff. & $\rfaithref$ \\
\midrule
ASL3DWord (gloss) & 100 & 339 & 874 & 113,708 & 0.752 \\
Phoenix-2014T & 50 & 325 & 4,399 & 6,500 & 0.874 \\
CSL-Daily & 1,000 & 2,731 & 2,504 & 54,620 & 0.673 \\
How2Sign ($\geq$5 w., cross-video) & 50 & 116 & 85 & 2,320 & 0.849 \\
\bottomrule
\end{tabular}
\end{table}

To interpret $\rfaith$ we need a reference value. It is not zero.
Two genuine human renditions of the \emph{same} sentence differ in
signer, speed, and style. For sentences that are signed by at least
two signers, we can measure
\begin{equation}
\rfaithref =
\frac{\mathbb{E}_{g,\, k \ne k'}\|z^*_{g_k} - z^*_{g_{k'}}\|}
   {\mathbb{E}_{g \ne g',\, k, k'}\|z^*_{g_k} - z^*_{g'_{k'}}\|},
\label{eq:rfaith_ref}
\end{equation}
where $g$ indexes same-text groups and $k$ renditions within a
group: the numerator is the distance between two renditions of the
same sentence, the denominator the distance across different
sentences. It shows how close a second human rendition comes to the
target, measured from each dataset's own ground truth. It is an
empirical reference, not a lower bound. A generator that outputs the
average of the valid renditions can fall below it without modeling
how signing varies.

\textbf{Because $\rfaithref$ differs across datasets, we propose no
absolute threshold for $\rfaith$.} A model far above its dataset's
$\rfaithref$ has collapsed. Reaching the reference is consistent with
faithful generation but does not prove it. \Cref{tab:reference} lists the measured references and the
pairs behind them; CSL-Daily, built from scripted cross-signer
re-performances, is the strictest. How2Sign's remaining exact repeats
are short greetings/fillers (``Hi!'', ``Okay?'') whose clips carry no
sentence-specific motion and would push the reference to
$\approx 1.0$, so we restrict to $\geq 5$-word cross-video repeats.

\subsection{Stability of the diagnostic}
\label{sec:diagnostic_robustness}
The numbers in \cref{tab:main} use the \emph{full} test split of each
dataset, so the reported values do not depend on which subsample of
test sentences is drawn. Even on the full test split, almost every
sentence-level model evaluated on How2Sign attains $\rfaith \approx 1$. A
fair concern is that this value reflects a problem with the metric
itself, not with the models: either (i) the MoAE cannot tell motions
apart in its latent space, or (ii) $\rfaith$ always returns
$\approx 1$ regardless of input. We rule both out with two checks,
and add a third on what the MoAE resolves.

\textbf{(1) MoAE initialization.} Re-training the MoAE from a
different random initialization (same architecture and data) and
re-running the diagnostic preserves the relative ordering of all
measured checkpoints; the diagnostic is robust to MoAE
training noise and not an artefact of one particular MoAE fit.

\textbf{(2) GT-vs-GT check: can $\rfaith$ ever move?}
To rule out (ii), we feed the diagnostic the strongest possible
``alignment'' case: a second GT rendition of the same sentence in
place of the model output. This yields the references of
\cref{sec:reference} ($0.67$--$0.87$), far below $1$, and the
healthy models of \cref{sec:coverage} reach them: the metric can
move; the collapsed models do not move it.

\textbf{(3) MoAE resolution on hands.} Per-joint-group reconstruction
error (normalized RMSE, the fraction of motion \emph{not} captured)
on full test splits: body/arms $0.17$--$0.20$, hands $0.30$--$0.37$
--- hands are harder but well encoded, so $\rfaith$ is sensitive to
the articulators carrying lexical content (facial expression is
outside the MoAE input; \cref{sec:conclusion}).

\section{Three Levels of Conditional Collapse}
\label{sec:three_layer}

The ratios decompose SLP failure into three independent
\emph{levels}:
\begin{description}[leftmargin=1.2em, itemsep=1pt, topsep=2pt]
\item[\Tone{} Initial pose.] Satisfied when the $t{=}0$ pose varies
   with the input text (\emph{varied}); failed when every generated
   initial pose is identical across sentences (\emph{frozen}).
   Visually inspectable in rendered output.
\item[\Ttwo{} Output diversity.] Satisfied when outputs vary across
   inputs as much as GT motions do; failed
   when outputs cluster around a near-identical mean motion
   ($\rdiv \to 0$).
\item[\Tthree{} Faithfulness.] Satisfied when each output aligns to
   its specific target and $\rfaith$ approaches the dataset's
   GT-vs-GT reference (\cref{sec:reference}); failed when alignment
   is no better than random ($\rfaith \approx 1$).
\end{description}

These levels are \emph{independently attainable}: a model can satisfy
\Tone and \Ttwo without attaining \Tthree. On How2Sign, the checkpoints we
rendered show two visual cases.
\textbf{(i) Frozen start $+$ jitter trajectory} (all three levels
collapsed): a near-identical canonical start regardless of input,
then small-amplitude noise around the GT marginal mean. This is the
dominant failure mode and includes the FID-best checkpoint (row~3 of
\cref{tab:main}).
\textbf{(ii) Varied start $+$ jitter trajectory} (\Tone, \Ttwo
satisfied, \Tthree collapsed): a visibly different start per input,
but the trajectory is still noise. It appears under aggressive
input narrowing (small vocabularies, short sentences).
The two full-corpus MDM checkpoints trained with the FK loss (rows~1
and~2 of \cref{tab:main}) satisfy \Ttwo{} numerically. Their outputs
vary across inputs more than GT motions do ($\rdiv \approx 1.8$), yet
each output is no closer to its own target than to a random one
($\rfaith$ $0.96$ to $0.98$).
\textbf{(iii) Varied start $+$ coherent motion} (all satisfied) is
never observed on How2Sign; it is reached on the dense-coverage
datasets of \cref{sec:coverage}.

\begin{figure}[t]
\centering
\includegraphics[width=0.76\linewidth,trim=0 4 0 0,clip]{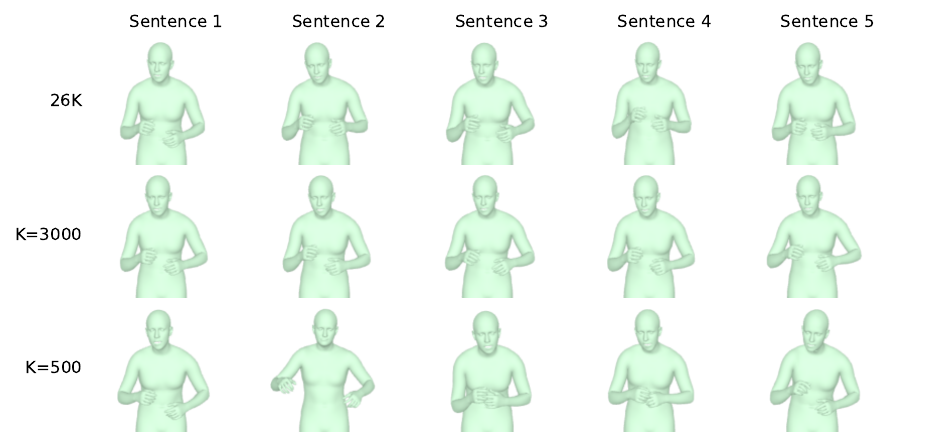}
\caption{Initial generated pose ($t = 0$) per checkpoint (rows);
each column is a different test sentence.
\textbf{Row 1:} full How2Sign $26$K. \textbf{Rows 2--3
($K{=}3000$, $K{=}500$):} training restricted to sentences within the
top-$3{,}000$ / top-$500$ vocabulary. 
Rows $1$--$2$ show
\emph{frozen} initial pose, identical regardless of the input sentence ($\tau_1$ collapse); row $3$ shows
\emph{varied} initial pose ($\tau_1$ satisfied). Variant definitions
in Appendix~A.}
\label{fig:initial_pose}
\end{figure}

\Cref{fig:initial_pose} shows the (i)-vs-(ii) contrast directly:
rows 1--2 emit a near-identical canonical start regardless of input
(\Tone{} collapsed), row 3 shows visibly different starts (\Tone{}
satisfied). Trajectory slices at five relative times
(Fig.~S1) show small-amplitude input-independent motion
where the corresponding GT rows show coherent gestures --- visual
\Tthree{} failure even for the strongest \Tone/\Ttwo{} checkpoints.

\section{Experiments}
\label{sec:experiments}

In all experiments, we refer to  ``checkpoint'' as one fully 
trained model with distinct design configurations (architecture, conditioning input,
dataset subset), trained from scratch and saved at the best epoch.

\textbf{Datasets.}
\textbf{How2Sign}~\cite{duarte2021how2sign} is an open-domain English
instructional ASL dataset: $31$K training / $2{,}357$ test sentences with
per-frame SMPL-X~\cite{pavlakos2019smplx} parameters; train
vocabulary $15{,}434$ words. 
Our default training pool keeps the $\sim$$26$K sentences of
$3$--$30$ words; a checkpoint trained on the unfiltered $31$K set
collapses identically (Table~S1).
\textbf{Phoenix-2014T}~\cite{forster2014phoenix} is a dataset of German weather
broadcasts comprised of $7{,}092$ train / $642$ test sentences
and SMPL-X fittings from the
Signs-as-Tokens release~\cite{zuo2025soke}.
\textbf{CSL-Daily}~\cite{zhou2021csldaily} is a Chinese daily-life
dataset comprised of $18{,}399$ train / $1{,}176$ test sentences with fittings,
from the same release~\cite{zuo2025soke}. Each training
sentence is performed by $2.8$ signers on average, and test
sentences are unseen. 
\textbf{ASL3DWord}~\cite{dong2024signavatar} is a dataset of isolated glosses with 
$1.2$K train / $339$ test clips over $103$ gloss types.
All four datasets use the same $53$-joint SMPL-X representation, so the
diagnostic pipeline 
remains unchanged.

\textbf{Conditioning inputs.}
Next we vary the type of conditioning input and its representation. 
\emph{Sentence}: the sentence is encoded into a single pooled
token by a frozen contrastive text encoder
CLIP-ViT-B/32~\cite{radford2021clip} for English and German, and
the Chinese-CLIP text tower~\cite{yang2022chineseclip} for Chinese (\cref{sec:modelfree}).
\emph{LLM keywords}: we prompt an open-source instruction-tuned LLM
(Qwen2.5-32B-Instruct~\cite{qwen2024qwen25} for English, Qwen2.5-14B-Instruct for German and Chinese) 
to reduce each sentence to a short list of representative content
words that will serve as {\em a pseudo-gloss}. This keeps the content
words (nouns, verbs, pronouns, negation, numerals, meaning-bearing
modals), drops function words, and lemmatizes the sentence. 
The details can be found 
in Appendix~B. 
A similar LLM extraction is used by~\citet{guo2025pgg} for sign
language translation. The keywords serve as the closest proxy gloss for How2Sign dataset that does not have gloss annotations; Phoenix-2014T and CSL-Daily provide \emph{native gloss} annotations,
letting us compare LLM keywords against linguist-annotated gloss.
\emph{Fused}: sentence and keywords are concatenated into a single CLIP input (``\texttt{sentence ; keywords}'').
\Cref{tab:kw_examples} shows per language examples of gloss and LLM keywords. Additional examples can be found in Table~S2. Note that LLM keywords keep source text order
while native gloss follows sign order.
All tested checkpoints vary only in the conditioning
\emph{input}. 

\begin{table}[t]
\centering
\scriptsize
\setlength{\tabcolsep}{3pt}
\caption{Example of keyword-extraction per language, with
the corpus's native gloss where one exists. Keywords keep source
order; native gloss follows sign order (CSL-Daily topic-fronts the
location; DGS uses indexing signs such as \texttt{IX}).}
\label{tab:kw_examples}
\begin{CJK}{UTF8}{gbsn}
\begin{tabular}{lp{0.36\linewidth}p{0.22\linewidth}p{0.28\linewidth}}
\toprule
 & \textbf{Sentence} & \textbf{Native gloss} & \textbf{LLM keywords} \\
\midrule
EN & You don't have to take big steps. & --- & you not take big step \\
DE & teilweise kr\"aftiger regen zieht heute nacht \"uber uns ostw\"arts & TEILWEISE REGEN KOMMEN IX & teilweise kr\"aftig regen ziehen heute nacht \"uber uns ostw\"arts \\
ZH & 我要去超市买椅子，你去吗？ & 超市 我 要 买 椅子 你 去 & 我 去 超市 买 椅子 你 去 \\
\bottomrule
\end{tabular}
\end{CJK}
\end{table}

\textbf{Models.}
We evaluate two model families that together cover the recurring
design choices in SLP: iterative diffusion and single-pass regression.
\textbf{(i) Diffusion.} Our main backbone is MDM~\cite{tevet2023mdm},
a Transformer denoiser that receives the noisy motion sequence
together with a timestep token and the frozen CLIP text token, and is
trained to predict the noise (architecture in Fig.~S2a).
It is trained with a per-frame pose MSE on the SMPL-X rotations plus a forward-kinematic
term $\mathcal{L}_{\mathrm{FK}}$ that penalizes $3$D joint-position
error after forward kinematics. One checkpoint (row 3 of
\cref{tab:main}) is trained with the pose MSE only; it attains the
best FID of all our How2Sign models and is included precisely because
FID ranks it first.
\textbf{(ii) Regression.} As a deliberately simple non-diffusion
baseline we train a clean Transformer-decoder regressor: $T$ learned
frame queries pass through a $6$-layer decoder whose cross-attention
memory is the single frozen CLIP token, and the network emits the
full motion sequence in one forward pass ($\sim$$26$M trainable
parameters; Fig.~S2b). It uses the same pose and FK
losses as MDM. Because it is deterministic,
$\rcond$ is not defined for it.
For all models the architecture, text encoder, and training schedule
are held fixed across data and conditioning variants, so differences
between rows reflect the data or the conditioning input rather than
tuning; hyperparameters are listed in Appendix~D.

\textbf{Evaluation.}
For each checkpoint we compute $(\rcond, \rdiv, \rfaith)$ and FID
on the full test split, render SMPL-X mesh videos, and visually
inspect initial pose and trajectory.

\subsection{Main negative result}
\label{sec:main_table}

We trained $14$ sentence-level How2Sign dataset checkpoints 
spanning two architecture families (diffusion, regression), two conditioning inputs (sentence, LLM keywords), while varying the four characteristics of the dataset: training-set
size ($13$K/$26$K/$31$K), vocabulary restriction (top-$3{,}000$ /
top-$500$ training vocabulary), utterance length ($\leq 3$ words;
single word), and loss composition. \Cref{tab:main} shows 
representative checkpoints. 
Every How2Sign checkpoint in \cref{tab:main} attains
$\rfaith \in [0.95, 1.00]$, far above its GT-vs-GT reference of
$0.849$, and no configuration attains \Tthree. Holding everything
else fixed, growing the training pool from $13$K to $26$K
sentences (rows 4--5) \emph{worsens} $\rdiv$ ($0.41 \to 0.34$)
while $\rfaith$ stays $\approx 0.95$, inside the collapse range.
Note that more data lets the model output the average motion more
confidently, yet it does not help the model to respond faithfully to
the input. \emph{Utterance length} (Table~S1). Decreasing $T = 100, 50, 32$ improves $\rdiv$ monotonically ($0.45 \to 0.46 \to 0.63$) leading to shorter sentences satisfying  \Ttwo. So \Tone{} and \Ttwo{} can be satisfactory
by narrowing the input enough (\cref{fig:initial_pose}: the
top-$500$-vocabulary regression checkpoint varies its initial pose),
but \Tthree{} cannot. 
\begin{table}[t]
\centering
\scriptsize
\setlength{\tabcolsep}{2.2pt}
\caption{Collapse diagnostics across the four datasets (ratios
defined in \cref{sec:diagnostic}; $\rfaith$ ref.\ $=$ the dataset's
GT-vs-GT reference, \cref{sec:reference}, healthy $=$ at or near
it). How2Sign: $5$ of $14$ checkpoints (all in Table~S1); the Phoenix-2014T and CSL-Daily blocks vary only the
conditioning string. 
$\rcond$: diffusion rows, two seeds per sentence;
regression rows are deterministic ($\rcond$ $=$ --); other `--' $=$
not computed.}
\label{tab:main}
\begin{tabular}{llrrrrr}
\toprule
\# & Model / conditioning & $N$ tr & FID & $\rcond$ & $\rdiv$ & $\rfaith$ \\
\midrule
\multicolumn{7}{@{}l}{\textbf{How2Sign} (open domain; $\rfaith$ ref.\ $0.849$)} \\
1 & MDM / sentence                  & 26K & 2.66 & 2.00 & 1.79 & 0.98 \\
2 & MDM / LLM keywords              & 26K & 2.57 & 2.86 & 1.76 & 0.96 \\
3 & MDM / sentence (FID-best)$^\dagger$ & 26K & \textbf{1.91} & 0.45 & \textbf{0.17} & \textbf{1.00} \\
4 & Regr.\ / sentence               & 13K & -- & -- & 0.41 & 0.96 \\
5 & Regr.\ / sentence               & 26K & -- & -- & 0.34 & 0.95 \\
\midrule
\multicolumn{7}{@{}l}{\textbf{Phoenix-2014T} (weather; $\rfaith$ ref.\ $0.874$)} \\
6  & MDM / sentence                 & 7K & 3.66 & 1.35 & 1.33 & 0.839 \\
7  & MDM / LLM keywords             & 7K & 2.14 & 1.66 & 1.34 & \textbf{0.809} \\
8  & MDM / native gloss             & 7K & 3.48 & 1.88 & 1.46 & 0.852 \\
9  & MDM / sent.\,$\oplus$\,keywords & 7K & 2.39 & 1.71 & 1.36 & 0.852 \\
10 & Regr.\ / sentence              & 7K & 1.39 & -- & 0.45 & 0.811 \\
\midrule
\multicolumn{7}{@{}l}{\textbf{CSL-Daily} (daily life; $\rfaith$ ref.\ $0.673$; Chinese-CLIP)} \\
11 & MDM / sentence                 & 18K & 2.35 & 1.55 & 1.15 & 0.798 \\
12 & MDM / LLM keywords             & 18K & 1.48 & 2.26 & 0.94 & 0.787 \\
13 & MDM / native gloss             & 18K & 1.63 & 2.22 & 1.20 & 0.788 \\
14 & MDM / sent.\,$\oplus$\,gloss   & 18K & 1.71 & 2.21 & 1.16 & \textbf{0.750} \\
15 & Regr.\ / sentence              & 18K & 1.34 & -- & 0.55 & 0.775 \\
\midrule
\multicolumn{7}{@{}l}{\textbf{ASL3DWord} (isolated gloss; $\rfaith$ ref.\ $0.752$)} \\
16 & MDM / gloss word               & 1.2K & -- & \textbf{4.80} & \textbf{1.14} & \textbf{0.731} \\
\bottomrule
\end{tabular}

\footnotesize{$^\dagger$Row 3 drops $\mathcal{L}_{\mathrm{FK}}$:
FID-best but marginal-collapsed. Row 16: same MDM backbone as rows
1--3; $0.731$ matches ASL3DWord's reference $0.752$.}
\end{table}

\subsection{The FID and \texorpdfstring{$\rfaith$}{faithfulness} disconnect}
\label{sec:fid_disconnect}

Across the $14$ How2Sign checkpoints 
and our NSA
re-implementation (\cref{sec:discussion}), FID spans from $1.91$ to
$151$ while $\rfaith$ spans only $0.89$--$1.00$. 
The best-FID model (row 3) is the \emph{most} collapsed
($\rfaith = 1.00$, $\rdiv = 0.17$) and its outputs cluster tightly
around the marginal mean, which is exactly what FID rewards.
The Phoenix-2014T and CSL-Daily blocks agree: their lowest FID
($1.39$, $1.34$) is the regression row, also lowest in $\rdiv$.
Hence FID alone does not properly account for text-to-motion
alignment. The full table can be found in Table~S1.

\subsection{BT-BLEU also fails to track faithfulness}
\label{sec:bt_disconnect}

We trained a motion-to-text back-translation (BT) model on How2Sign
without pretraining. Without pretraining, How2Sign alone does not
train a usable translation network. The video translator
of~\citet{youtubeasl2023} reaches only BLEU-4 $\approx 1.22$ in this
regime. Our BT model reaches BLEU-4 $\approx 2.5$ even on
ground-truth motion, and $1.24$ on our generated motion.
Its translation predictions are
often input-independent template language unrelated to the GT sentence
(see \cref{tab:bt_examples}). The non-zero BLEU score is due to n-gram overlap with
high-frequency vocabulary where an autoregressive decoder in the back-translation model is estimating a
near-marginal prior. BT-BLEU on How2Sign therefore reflects the
weakness of the evaluator more than the quality of the generator.
$\rfaith$ captures this failure directly. BT-BLEU does not.
\begin{table}[t]
\centering
\scriptsize
\setlength{\tabcolsep}{3pt}
\caption{Back-translation predictions for our generated motion on
the How2Sign test split (BLEU-4 $= 1.24$, no pretraining). They are
input-independent instructional template language or degenerate
repetition, unrelated to the ground truth. Evaluator details in Appendix~C.}
\label{tab:bt_examples}
\begin{tabular}{p{0.40\linewidth} p{0.54\linewidth}}
\toprule
\textbf{Ground truth} & \textbf{Back-translation prediction} \\
\midrule
Boom, just like that. &
so, i'm going to take my right hand and i'm going to take my left
hand and i'm going to take my left hand to my left hand. \\[0.2em]
Now it's your responsibility to provide the healthy food. &
you can do it on both sides of the legs. \\[0.2em]
Depending on the focus, if it's a high-intensity workout$\ldots$ &
you can have a very, very light, and you can have a very light, very
light, very light, v$\ldots$ \\
\bottomrule
\end{tabular}
\end{table}

\section{The Coverage Argument}
\label{sec:coverage}
In the following section we present a controlled experiment that sheds light on model collapse when training on How2Sign dataset and performance on the other datasets.  
We use the \emph{same} two architectures, 
frozen text encoder, and diagnostic on four datasets that
differ in how densely their training sentences cover the conditions their test sets draw from. Each dataset uses its own MoAE and its
own GT-vs-GT reference. The diagnostic ratios are dimensionless and the values are compared across datasets only by \emph{band
membership}, at-reference vs.\ far-above-reference values. 

\subsection{Same models, four datasets}
\label{sec:spectrum}

\begin{figure}[t]
\centering
\includegraphics[width=0.84\linewidth,trim=0 0 0 106,clip]{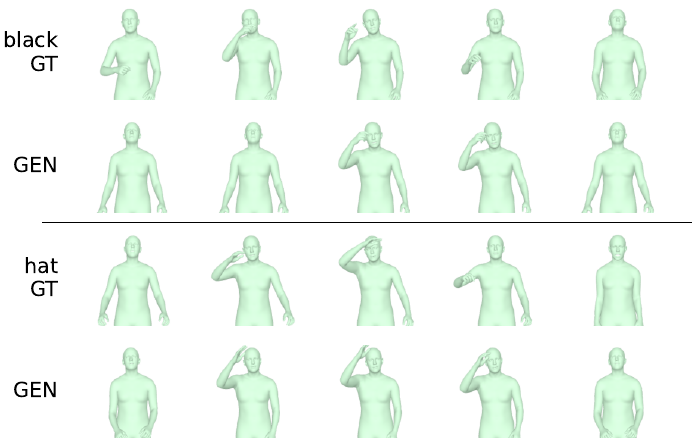}
\caption{ASL3DWord test-set renders for two glosses (\textbf{black},
\textbf{hat}): GT on top, generation below, $5$ frames per row
sampled at quantiles of GT cumulative motion energy (upper body).
The same MDM backbone that collapses on How2Sign produces the
target sign here (row 16 of \cref{tab:main}).}
\label{fig:asl3dword}
\end{figure}

\begin{figure}[t]
\centering
\includegraphics[width=0.742\linewidth]{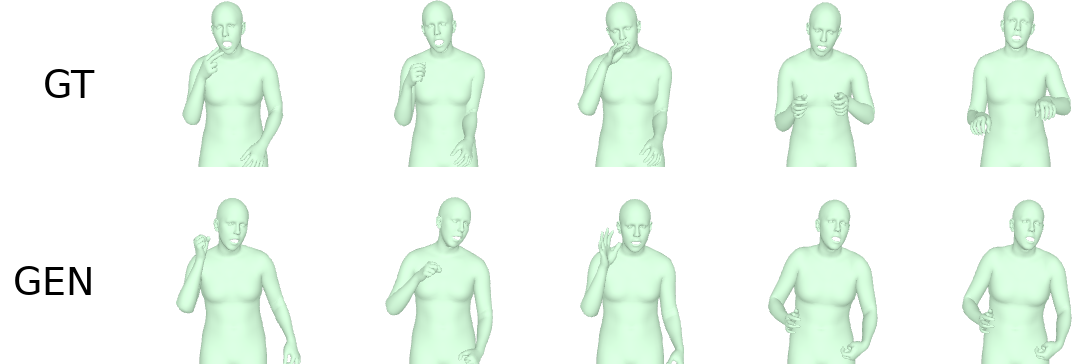}\\
{\footnotesize (a) Phoenix-2014T, recurring broadcast opener: $0.66$ (human band $0.74$)}\\[3pt]
\includegraphics[width=0.81\linewidth]{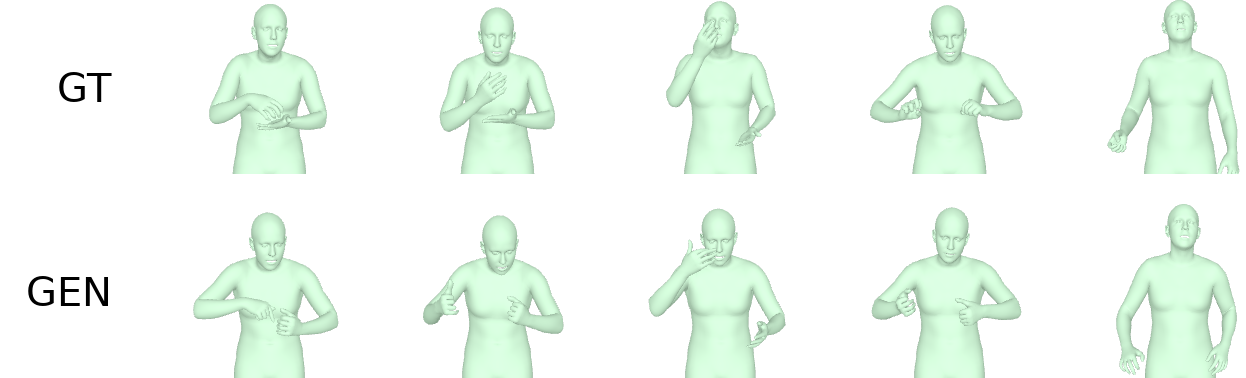}\\
{\footnotesize (b) CSL-Daily, unseen sentence: $0.70$ (human band $0.59$)}
\caption{Best-aligned rendered test-set generation per dataset (GT
top, generation bottom) as \emph{DTW-matched} frame pairs, with
DTW$/$random per example (\cref{sec:framelevel}). (a) \emph{``und nun
die wettervorhersage\ldots''}, inside the human--human band;
(b) an unseen CSL-Daily sentence (Chinese original in the
Appendix~G), which reaches the generation p10 but not the
stricter CSL-Daily human band. Further examples in Fig.~S3.}
\label{fig:qual_examples}
\end{figure}

The main results on Phoenix-2014T, CSL-Daily, and ASL3DWord are reported in~\cref{tab:main}.
On How2Sign, both architectures have a $\rfaith$ gap of
$0.10$--$0.13$ above the $0.849$ reference. On Phoenix-2014T both models reach $\rfaith$ \emph{below} the $0.874$
reference. The generated motion is as close to
its target as another human rendition of the same sentence. All four conditioning-string variants stay in that
range (\cref{sec:conditioning}); \cref{fig:qual_examples}a shows a
rendered generation against ground truth.
CSL-Daily's reference ($0.673$) is the
strictest of the four, with training sentences performed by $2.8$
signers on average. 
Using the Chinese-CLIP text embedding~\cite{yang2022chineseclip} (\cref{sec:modelfree}), all five
models attain $\rfaith$ $0.75$--$0.80$ (\cref{fig:qual_examples}b).
On gloss-level ASL3DWord the same MDM backbone attains  $\rfaith$ of $0.731$
against a $0.752$ reference (\cref{fig:asl3dword}).
The \emph{largest} corpus
(How2Sign, $31$K) is the one that fails its reference, while
$4\times$ smaller Phoenix-2014T and $26\times$ smaller ASL3DWord approach
their reference. This corroborates the observation that \textbf{more paired data does not produce faithfulness, and
less does not preclude it}. It is the structure of the dataset that is the determining factor of the performance.  

\subsection{Frame-level alignment}
\label{sec:framelevel}

In \cref{fig:qual_examples} the generated frames visually track
their GT. To check whether this holds beyond selected examples, we
measure frame-level agreement directly, complementing the MoAE
ratios, which represent each motion as a single embedding vector.
We compare motions pose by pose with dynamic time warping (DTW):
we convert each motion into per-frame $3$D coordinates of the $44$
upper-body joints, align the two sequences in time, and sum the
per-frame distances. This sum is divided by the same quantity
computed for pairs of \emph{unrelated} sentences.

We score two kinds of pairs. We take $40$ sentences per dataset
that were performed at least twice and score the two human
renditions against each other: $0.74$ on Phoenix-2014T and $0.59$
on CSL-Daily. Note that the two renditions come from different
signers. For the models, we score the generated motion against the
GT of the same test sentence ($40$ sentences per dataset): $0.86$
and $0.76$, farther from the GT than two humans are from each
other. Only the most repeated sentences close this gap
(\cref{fig:qual_examples}a). In short, passing the MoAE test does
not yet mean human-level frame-by-frame agreement: our ``reaches
its reference'' claims are about whole sequences (full protocol in
Appendix~G).

\subsection{Model-free evidence: dispersion and retrieval}
\label{sec:modelfree}

If the determining factor of collapse is conditional coverage, it should be measurable in
the conditioning space itself, with no generator involved. We test
this two ways.

\begin{table}[t]
\centering
\small
\setlength{\tabcolsep}{3pt}
\caption{Conditioning-space density (cosine similarity; higher $=$
denser): English CLIP for How2Sign and Phoenix; Chinese-CLIP for
CSL-Daily, since English CLIP barely separates Chinese sentences
(Appendix~F). Mean-centered $=$ train centroid
subtracted, removing encoder anisotropy.}
\label{tab:dispersion}
\begin{tabular}{lccc}
\toprule
Text-embedding cosine & How2Sign & Phoenix & CSL-Daily$_{\text{zh}}$ \\
\midrule
within-train pairwise            & 0.45 & 0.73 & 0.72 \\
test $\to$ nearest-train         & 0.80 & 0.95 & 0.92 \\
\;\;same, mean-centered            & 0.63 & 0.79 & 0.71 \\
\bottomrule
\end{tabular}
\end{table}

\paragraph{Conditioning space dispersion.}
On the frozen CLIP text encoder the generators condition on, we
measure (a) \emph{within-train dispersion}: pairwise cosine among
training sentences, and (b) \emph{coverage}: each test sentence's
cosine to its nearest training neighbor (\cref{tab:dispersion}).
Phoenix is far denser than How2Sign on every measure. Its 
$5$th-percentile coverage ($0.89$) exceeds How2Sign's \emph{median}
($0.80$); mean-centering preserves the gap ($0.79$ vs $0.63$), and a natively multilingual XLM-R
encoder~\cite{conneau2020xlmr} preserves its direction, so the gap
is not an artifact of an English encoder reading German.

CSL-Daily is measured on the Chinese-CLIP text
tower~\cite{yang2022chineseclip}; its centered coverage ($0.71$)
lands between How2Sign ($0.63$) and Phoenix ($0.79$), matching its
in-between model results.

\paragraph{Nearest-neighbor retrieval sweep.}
For each test sentence, retrieve the nearest training sentence by
text embedding and use its GT \emph{motion} as the prediction --- an
architecture-free ``model'' whose faithfulness depends only on
coverage (full curves in Appendix~F). On How2Sign, $\rfaith$ stays flat at
$\sim$$0.98$ as the pool grows from $500$ to $26{,}000$ sentences
($50\times$): even the nearest of $26$K training motions is, on
average, no closer to a test target than a random one. On Phoenix
the same sweep \emph{improves} with pool size ($0.905 \to 0.867$),
reaching the reference ($0.874$); on CSL-Daily (Chinese-CLIP) it
improves ($0.888 \to 0.868$) but stays far above the $0.673$
reference: test sentences are unseen, so no pool offers
a rendition to copy --- and every CSL-Daily generator
lands \emph{below} this $0.868$ bound (rows 11--15 of
\cref{tab:main}), composing beyond retrieval. On How2Sign, more in-domain data adds no
coverage, so no model trained on it should be expected to condition
--- and none does.

\subsection{Text encoder effects}
\label{sec:controls}

\textbf{Same encoder, opposite outcomes.} The same frozen
English-pretrained CLIP text encoder reads \emph{German} on Phoenix-2014T and the model is faithful ($\rfaith = 0.839$) while encoding 
English on How2Sign collapses ($\rfaith = 0.98$). If weak text representations were the primary cause, this ordering would be reversed.\\
\noindent
\textbf{Encoder swap.} A more powerful frozen T5-base text encoder~\cite{raffel2020t5} in the
$26$K How2Sign regression 
leaves $\rho_{\mathrm{faith}}$ essentially unchanged ($0.954$ with CLIP, $0.942$ with T5). 

\noindent
\textbf{Text shuffling.} On How2Sign's unfiltered $31$K
sentence-MDM checkpoint (row 6 of Table~S1), replacing
the input sentence moves the output $1.05\times$, no more than re-sampling the
noise seed. Word-shuffling leaves
faithfulness almost unchanged ($0.963 \to 0.970$), which suggests that the model 
output is not affected by conditioning text. This is in contrast to Phoenix-2014T, where 
sentence replacement moves the output well beyond
noise ($1.35\times$) while word-shuffling moves it less than a new
seed ($0.49\times$).  
$\rfaith$ stays near the reference ($0.838 \to 0.855$): the model reacts to the words themselves, not their order.

\subsection{Coverage is necessary, not sufficient}
\label{sec:necessary}

We retrained Progressive Transformers~\cite{saunders2020progressive},
a representative of the autoregressive architecture family
on the same clean Phoenix-2014T SMPL-X data. 
During inference (step counter
and output length governed by the model's own predictions rather
than supplied from ground truth), it
\textbf{collapses} ($\rfaith = 0.9999$; output spread $0.025\times$
GT) 
(see more details in Appendix~H).

To conclude: with sparse coverage every architecture we tested 
collapses, and with dense coverage the architecture and training
protocol still decide the outcome. The diagnostic exposes both
kinds of failure.

\subsection{Conditioning representation}
\label{sec:conditioning}

On the dense-coverage corpora we study the effect of the conditioning
representation, considering four options, sentence encoding, 
LLM-extracted keywords (\cref{sec:experiments}), native gloss
(if available in the annotations), and a fused input. Native gloss
is the annotation upper bound for the LLM keywords: their gap on
annotated corpora measures the proxy's quality where no annotation
exists (How2Sign). On How2Sign, using LLM keywords does not avoid
collapse (row 2 of \cref{tab:main}). 

On Phoenix, all four alternatives stay at or near the $0.874$ reference
(rows 6--9): once coverage is dense, no representation choice
re-introduces collapse. LLM keywords give the best faithfulness
($0.809$), matching or exceeding the linguist-annotated gloss
($0.852$). This is relevant for corpora like How2Sign that do not 
have gloss annotations. 
On CSL-Daily (Chinese-CLIP), all four cases share similar range 
($0.75$--$0.80$; rows 11--15  of \cref{tab:main}): LLM keywords ($0.787$) and
native gloss ($0.788$) are indistinguishable, so token
\emph{order} does not decide the outcome, and the fused
sentence$\oplus$gloss input is best ($0.750$). 
These differences are small (single runs); the
robust conclusion is that no conditioning text turns a collapsing
dataset into a healthy one, or the reverse.

\section{Discussion}
\label{sec:discussion}

\paragraph{Reproducibility note on Neural Sign Actors.}
NSA~\cite{baltatzis2023nsa} reports the lowest sentence-level
How2Sign FID in the literature but releases neither code nor
weights. Following the published description, our re-implementation
needed several debugging rounds to converge; our best run reaches
FID $\approx 151$ with $\rfaith \approx 1.00$ and no input-specific
motion in rendered samples. We do not interpret this discrepancy;
it does not affect our claim, since even an NSA-class FID is
consistent with \Tone/\Ttwo{} satisfied $+$ \Tthree{} collapsed.
That is the best regime any of our How2Sign checkpoints reaches. We read prior FID
gains in this line as improvements in \Tone{} and partial \Ttwo,
not \Tthree.
\paragraph{Reported successes on Phoenix and CSL-Daily.}
Qualitatively convincing generation is reported on
Phoenix-2014T~\cite{saunders2020progressive,huang2021fast,xie2024g2pddm,yin2024t2sgpt}
and CSL-Daily~\cite{zuo2025soke,zuo2024spoken2sign}; that evidence
is structurally consistent with our frame-level analysis. The
displayed sentences concentrate on the densest conditions ---
Progressive Transformers leads with the canonical broadcast opener,
three of SOKE's four Phoenix examples are broadcast greetings ---
precisely the conditions our DTW analysis places inside the human
band \emph{for our models as well} (\cref{sec:framelevel}). Every
figure shows one GT rendition and no human--human baseline, in formats that hide frame-level error, and
no verifiable generator weights were obtainable. Whether any attains
human-band frame alignment is thus open --- and directly testable
with our released MoAE, references, and DTW protocol. This is not a
claim that these systems fail; it supplies the missing calibration.

\paragraph{Recommended evaluation protocol.}
SLP papers should report $(\rcond, \rdiv, \rfaith)$ alongside FID
and BT-BLEU, judging $\rfaith$ against the dataset's GT-vs-GT
reference (\cref{eq:rfaith_ref}) on a \emph{shared} released MoAE,
and, where same-sentence GT pairs exist, frame-level DTW
alignability against the human--human baseline
(\cref{sec:framelevel}). $\rfaith$ is a \emph{necessary-condition}
check: passing it does not certify linguistic adequacy, which
requires native-signer evaluation.

\paragraph{What would fix How2Sign?}
Within-corpus scaling does not add coverage (\cref{sec:modelfree}),
so the candidates are external: large-scale text-to-motion
pretraining, transfer from dense-coverage corpora, or word-level
supervision injected from outside the paired corpus --- the route
SOKE's dictionary retrieval takes~\cite{zuo2025soke}, consistent with
the coverage account. The sign-to-text direction points the same
way: a translator trained directly on How2Sign reaches only BLEU-4
$\approx 1.22$, with faithful translation emerging only after $11$K
hours of pretraining~\cite{youtubeasl2023}. None is evaluated under
our diagnostic here; pretraining remains plausible but undemonstrated.

\paragraph{Societal impact.}
SLP evaluation rigor directly affects Deaf and hard-of-hearing
users: metrics that reward text-independent motion can promote
systems producing fluent-looking but meaningless signing. A
diagnostic calibrated against real signers' renditions is a
prerequisite for trustworthy progress claims in accessibility.

\section{Conclusion}
\label{sec:conclusion}

We introduced a training-free diagnostic that exposes conditional
collapse in sentence-level SLP, judged against per-dataset GT-vs-GT
references, and showed that (i) $14$ How2Sign checkpoints across
two architecture families never attain faithfulness while FID
varies $\sim$$80\times$ uncorrelated with it, and (ii) the driver
is per-condition coverage: the same architectures reach their
sequence-level references on densely-covered Phoenix-2014T and
ASL3DWord with less paired data, condition partially on CSL-Daily,
and collapse on How2Sign at every available scale. A frame-level
DTW lens calibrated on human--human consistency is stricter still:
even at-reference Phoenix reaches the human band only on its densest
formulaic conditions. Coverage is necessary but not sufficient ---
Progressive Transformers collapses on dense Phoenix --- so the
diagnostic separates data-limited from architecture-limited
failure. We release the MoAE checkpoints and
diagnostic code under permissive licenses.

\paragraph{Limitations.}
Our checkpoints do not cover the full design space; large-scale
text-to-motion pretraining remains untested. The MoAE
omits facial expression and non-manual markers, and hands are
reconstructed less precisely than body ($\mathrm{nRMSE}$
$0.30$--$0.37$ vs $0.17$--$0.20$): $\rfaith$ is a
necessary-condition check on manual articulation, not a certificate
of linguistic adequacy. Correlating the diagnostic with
native-signer judgment is the key next step.

{
    \small
    \bibliographystyle{ieeenat_fullname}
    \bibliography{references}
}

\newif\ifappendix
\appendixtrue
\ifappendix
\clearpage
\setcounter{section}{0}
\setcounter{figure}{0}
\setcounter{table}{0}
\setcounter{equation}{0}
\renewcommand{\thesection}{\Alph{section}}
\renewcommand{\thefigure}{S\arabic{figure}}
\renewcommand{\thetable}{S\arabic{table}}
\renewcommand{\theequation}{S\arabic{equation}}
\twocolumn[\centering\Large\textbf{Appendix}\vspace{1.0em}]
\noindent This appendix contains the full How2Sign result table with the initial-pose and
trajectory figures (\cref{sec:supp_h2s}); the LLM keyword extraction
prompts (English, German, Chinese) with uncurated examples
(\cref{sec:supp_kw}); the back-translation evaluator and its predictions
(\cref{sec:supp_bt}); implementation details of the MoAE, the diffusion
and regression models, and the training subsets (\cref{sec:supp_impl});
the per-dataset GT-vs-GT reference protocol (\cref{sec:supp_ref}); the
model-free coverage analyses and controls (\cref{sec:supp_cov}); the
frame-level DTW protocol with additional matched examples
(\cref{sec:supp_dtw}); and the code-level audit of Progressive
Transformers (\cref{sec:supp_pt}). Sections, tables, and figures prefixed by a letter
or ``S'' belong to this appendix.

\section{Full How2Sign results}
\label{sec:supp_h2s}

\Cref{tab:supp_main} reports all $14$ sentence-level How2Sign
checkpoints (rows~1--14), the How2Sign GT-vs-GT reference (row~15), and
the gloss-level ASL3DWord checkpoint (row~16). Rows 1--5 and 16 of Table~3 are rows 1, 2, 5, 7, 8,
and 16 of this table.

\begin{table*}[t]
\centering
\small
\setlength{\tabcolsep}{4pt}
\caption{Conditional-collapse diagnostics on How2Sign
(rows~1--14 plus GT reference row~15) and ASL3DWord
(row~16; double rule = different MoAE).
$\rcond$: signal/noise ratio. $\rdiv$: output/GT diversity ($\to 1$).
$\rfaith$: target alignment, judged against each dataset's GT-vs-GT
reference (How2Sign $0.849$; ASL3DWord $0.752$; \cref{sec:supp_ref}).
FID: diffusion only. \emph{Visual}: \textbf{F}/\textbf{V} =
frozen/varied initial pose; \textbf{J}/\textbf{M} = jitter/coherent
motion. \textbf{None of the $14$ How2Sign checkpoints reach the
How2Sign reference.}}
\label{tab:supp_main}
\begin{tabular}{lllrrrrrrc}
\toprule
\# & Family & Variant & $N$ train & $T$ &
FID $\downarrow$ & $\rcond$ & $\rdiv$ & $\rfaith$ & Visual \\
\midrule
1 & Diffusion & MDM + sentence                           & 26K & 100 & 2.66 & 2.00 & 1.79 & 0.98 & F+J \\
2 & Diffusion & MDM + LLM keywords                       & 26K & 100 & 2.57 & 2.86 & 1.76 & 0.96 & F+J \\
3 & Diffusion & MSTrans + sentence                       & 26K & 100 & 2.52 & 2.66 & 1.63 & 0.96 & F+J \\
4 & Diffusion & MSTrans + LLM keywords                   & 26K & 100 & 2.68 & 2.55 & 1.90 & 0.96 & F+J \\
5 & Diffusion$^\dagger$ & MDM + sentence (FID-best)$^\dagger$ & 26K & 100 & \textbf{1.91} & 0.45 & \textbf{0.17} & \textbf{1.00} & F+J \\
6 & Diffusion & MDM + sentence (unfiltered)              & 31K & 100 & 2.09 & 1.07 & 1.11 & 0.98 & F+J \\
\midrule
7 & Regression & 13K random sent                          & 13K & 100 & -- & -- & 0.41 & 0.96 & F+J \\
8 & Regression & 26K sent                                 & 26K & 100 & -- & -- & 0.34 & 0.95 & F+J \\
9 & Regression & K=3000 sent                              & 13.4K & 100 & -- & -- & 0.24 & 0.97 & F+J \\
10 & Regression & K=500 sent                              & 2.9K & 100 & -- & -- & 0.45 & 0.95 & V+J \\
11 & Regression & K=500 keywords                          & 2.9K & 100 & -- & -- & 0.48 & \textbf{0.92} & V+J \\
12 & Regression & K=3000 keywords                         & 13.4K & 100 & -- & -- & 0.28 & 0.96 & F+J \\
13 & Regression & short3 ($\le 3$-word)                   & 1.5K & 50 & -- & -- & 0.46 & 0.97 & V+J \\
14 & Regression & short1 ($1$-word)                       & 564 & 32 & -- & -- & 0.63 & \textbf{0.89} & V+J \\
\midrule
15 & GT motions$^\ddagger$ & cross-video same-sentence reference   & --- & --- & -- & -- & 1.00 & 0.849 & V+M \\
\midrule\midrule
16 & Diffusion$^\|$ & MDM (gloss-level, ASL3DWord)         & 1.2K & 60 & -- & \textbf{4.80} & \textbf{1.14} & \textbf{0.73} & V+M \\
\bottomrule
\end{tabular}

\footnotesize{Diffusion rows 1--4: default
$\mathcal{L}_{\mathrm{pose}}+\mathcal{L}_{\mathrm{FK}}$. Regression
rows are deterministic, so FID/$\rcond$ are uninformative (``--'').
$^\dagger$Row 5 drops $\mathcal{L}_{\mathrm{FK}}$: FID-best but
marginal-collapsed. $^\ddagger$Row 15: How2Sign GT-vs-GT reference
(\cref{sec:supp_ref}), i.e.\ the value a second human rendition of the
same sentence attains.
$^\|$Row 16: same MDM backbone and loss on ASL3DWord;
$\rfaith{=}0.73$ matches the within-gloss reference
$\rfaithref{=}0.752$.}
\end{table*}

\paragraph{Input modality} (\texttt{sent} / \texttt{keywords}).
\emph{Sentence}: the raw English How2Sign sentence, encoded by
frozen CLIP-ViT-B/32~\cite{radford2021clip}.
\emph{LLM keywords}: we prompt an open-source LLM to extract
\emph{pseudo-glosses} (short ordered lists of base-form content
words) from sentences; the exact prompt, model, and representative
outputs are in \cref{sec:supp_kw}. ASL gloss annotations are
unavailable for How2Sign, so the LLM keywords serve as the closest
available proxy.

\paragraph{Training subset.} The unfiltered How2Sign release has
$31$K training sentences. Our default training pool is the
$\sim$$26$K subset obtained by a $3$-to-$30$ word sentence-length
filter that removes long-tail outliers; rows~1--5 and rows~8--14 of
\cref{tab:supp_main} use this $26$K pool. As an end-to-end check
that the filter is not load-bearing, we also trained an identical
MDM+sentence variant on the unfiltered $31$K-sentence release
(row~6): $\rfaith = 0.98$ matches the $26$K run exactly, and the
other diagnostic ratios fall in the same band. Beyond these two,
we evaluate a $13$K random subsample of the $26$K pool (row~7) and
two families of input-narrowing variants:
\emph{vocabulary-restricted} (\texttt{K=K}): sentences whose words
all fall within the top-$K$ most-frequent training vocabulary
(\texttt{K=3000}, \texttt{K=500}); and \emph{short-sentence}:
\texttt{short1} (single-word) or \texttt{short3} ($\le 3$-word)
sentences drawn from the full $31$K-sentence release (these
bypass the $3$-to-$30$ word filter by design). Exact train/test
counts are in \cref{sec:supp_impl}. Row labels combine the two axes
(e.g.\ \texttt{K=500 keywords}). Each filter is applied to the raw
English sentence regardless of encoder input, so corresponding
\texttt{sent} and \texttt{keywords} rows use identical training and
test sentences and differ only in what the encoder sees.

\paragraph{Training and evaluation.}
Within each model family, the same training settings are used across
all data variants: batch size $256$, learning rate $2 \times 10^{-4}$,
and $100$ epochs for sentence-length variants; batch size $64$ and
$200$ epochs for short-sentence variants. For each checkpoint we
compute $(\rcond, \rdiv, \rfaith)$ and FID on the full test split.
When training was vocabulary-restricted, we evaluate on the matching
test subset (the test sentences whose words also pass the filter). We
additionally render generated motions as SMPL-X mesh videos and
visually inspect them for \Tone{} and trajectory coherence.

\paragraph{One axis at a time.}
Holding everything else fixed:
\emph{quantity} (rows 7 vs 8): $13$K $\to$ $26$K \emph{worsens} $\rdiv$
($0.41 \to 0.34$); $\rfaith$ stays in the collapse band (both
$\approx 0.95$). More data lets the model output the average motion
more confidently; it does not teach the model to respond to the input.
\emph{Vocabulary restriction} (rows 8, 9, 10): shrinking the training
vocabulary from $15{,}434$ to $3{,}000$ and $500$ types leaves
$\rfaith$ at $0.95$--$0.97$; only the initial pose becomes varied at
$K{=}500$ (V+J). \emph{Utterance length} (rows 10, 13, 14):
$T = 100, 50, 32$ improves $\rdiv$ monotonically
($0.45 \to 0.46 \to 0.63$); shorter sentences satisfy \Ttwo.
\emph{Backbone} (rows 1--4): the multi-scale MSTrans denoiser
(\cref{sec:supp_impl}) changes FID by $<0.2$ and $\rfaith$ by $\le 0.02$.
So \Tone{} and \Ttwo{} can be satisfied by narrowing the input
enough, but \Tthree{} cannot: changing data quantity, vocabulary
density, sequence length, or backbone all leave $\rfaith$ far above the
$0.849$ reference.

\paragraph{FID range.} Across rows~1--14 plus our Neural Sign Actors
re-implementation (Sec.~7; FID $\approx 151$, the upper
end), FID spans $1.91$ to $151$ while $\rfaith$ spans only $0.89$ to $1.00$; the Pearson
correlation between them is near zero.

\begin{figure}[t]
\centering
\includegraphics[width=\linewidth]{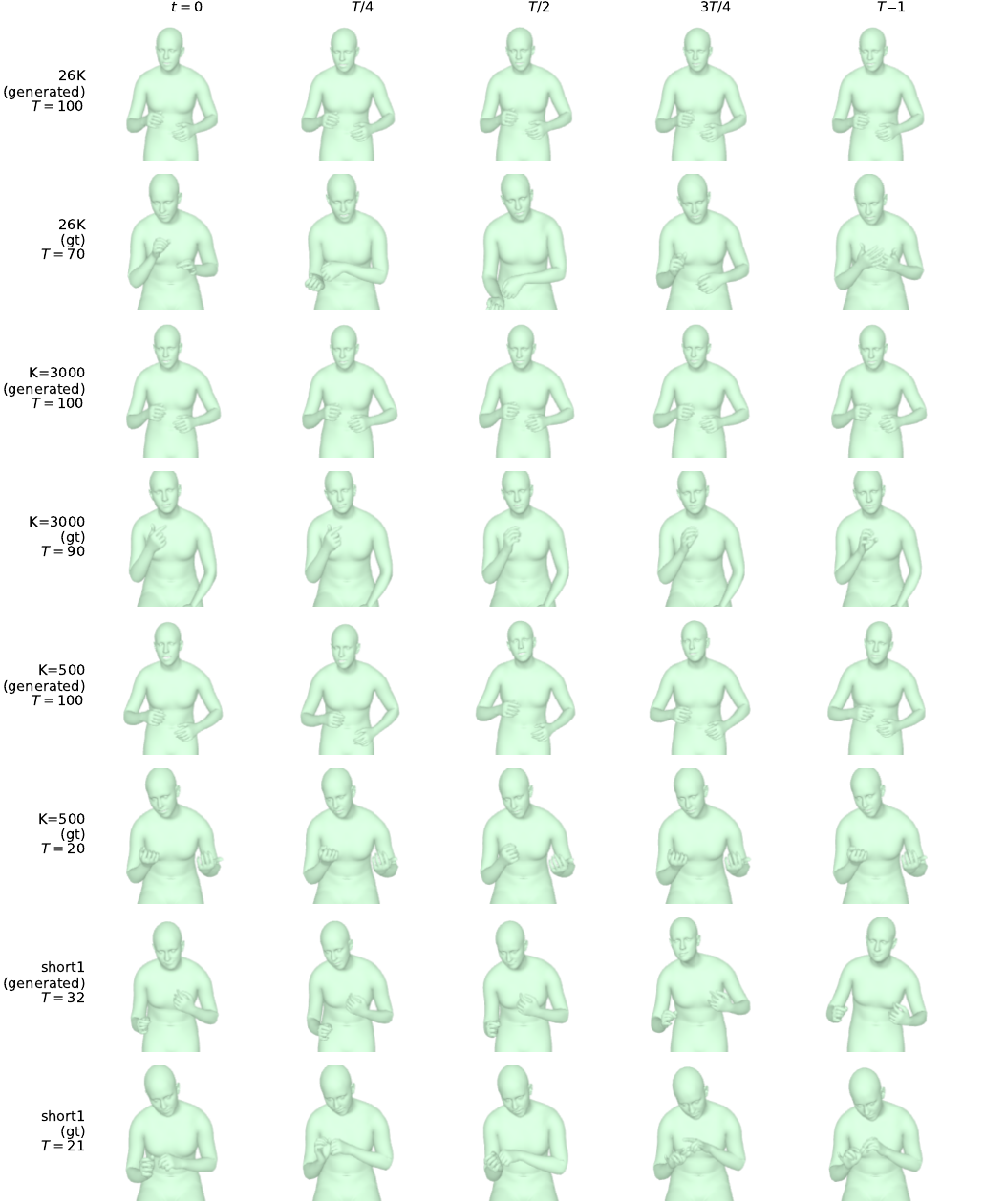}
\caption{Trajectory slices at relative times
$t \in \{0, T/4, T/2, 3T/4, T{-}1\}$ for the three checkpoints of
Fig.~1 plus the single-word \texttt{short1} checkpoint
(row 14 of \cref{tab:supp_main}): generated motion (top row of each pair)
above its corresponding ground truth (bottom row). $T$ differs per
row (shown in the row label) and \emph{differs between the generated
and GT rows of the same pair}: the columns share a relative phase,
not an absolute frame index. Each row pair also uses a \emph{different
test sentence} (one sample per checkpoint, sentences are not aligned
across checkpoints). The generated trajectories exhibit small-amplitude
jitter independent of the input, whereas the GT trajectories show
coherent gestures: visual confirmation of \Tthree{} failure across all
checkpoints, including the most strongly-satisfied \Tone/\Ttwo{} ones.}
\label{fig:supp_trajectory}
\end{figure}

\paragraph{Gloss-level ASL3DWord (row 16).}
We apply the exact same MDM backbone, loss, and diagnostic pipeline
to ASL3DWord~\cite{dong2024signavatar} (WLASL-101 SMPL-X subset,
$1.2$K train / $339$ test / $103$ glosses, $T{=}60$), where the task
is one English gloss $\to$ one isolated sign. Row 16 reports
$\rcond = 4.80$, $\rdiv = 1.14$, $\rfaith = 0.73$ on the full test
split; $\rfaith = 0.73$ matches the ASL3DWord GT-vs-GT same-gloss
reference $\rfaithref = 0.752$ (\cref{sec:supp_ref}). WLASL records
each word with multiple signers, and the same word performed by
different signers varies substantially in hand shape and motion, so
even a perfect generator that copies a real recording of the target
word cannot score below this within-word across-signer reference; the
checkpoint lands right at it (renders in Fig.~2). This positive example is not a new
word-level SLP method; it confirms that the diagnostic tells working
models apart from collapsed ones.

\section{LLM keyword extraction}
\label{sec:supp_kw}

The \emph{LLM keywords} conditioning input (Sec.~5) is
produced by prompting an open-source instruction-tuned LLM to reduce
each sentence to a short ordered list of base-form content words
(a pseudo-gloss). For English (How2Sign) we use
Qwen2.5-32B-Instruct~\cite{qwen2024qwen25} with greedy decoding and
\texttt{max\_new\_tokens}~$=256$; for German (Phoenix-2014T) and
Chinese (CSL-Daily) we use Qwen2.5-14B-Instruct with greedy decoding
and \texttt{max\_new\_tokens}~$=96$. No system message is used beyond
the prompt itself. Keywords are extracted once per sentence and cached;
every training variant of a corpus reads the same cache, so the
sentence, keyword, and fused variants of one corpus train on the
identical sample set (the length filter is always computed on the
original sentence). The prompts share one three-step recipe
(phrase-level chunking $\to$ keep/drop decision $\to$ citation form),
adapted per language: the German prompt handles auxiliary chains,
separable/particle verbs, and case inflection; the Chinese prompt
handles aspect particles, measure words, and the
\begin{CJK}{UTF8}{gbsn}把/被\end{CJK} constructions, and requires
standard word segmentation. All three prompts instruct the model to
keep source order; the prompts are reproduced below with
instructions verbatim and the few-shot example lists abridged to
four per language.

\begin{table*}[t]
\centering
\small
\setlength{\tabcolsep}{5pt}
\caption{Uncurated keyword-extraction examples in the three languages
(six random How2Sign training sentences; two sentences each for
Phoenix-2014T and CSL-Daily), with the corpus's native gloss where one
exists. English: function words, copulas, and auxiliary chains are
dropped and the remaining tokens lemmatised. LLM keywords keep the
source sentence order, whereas native gloss follows sign order (the
CSL-Daily gloss topic-fronts the location
\begin{CJK}{UTF8}{gbsn}超市\end{CJK}; the Phoenix DGS gloss uses
indexing and compound signs such as \texttt{IX}, \texttt{WIE-AUSSEHEN}
that no text-only extractor can produce). The second German row shows
a lemmatization slip (\emph{landeteil}); such artefacts are rare and
systematic, and do not change any verdict.}
\label{tab:supp_kw_examples}
\begin{CJK}{UTF8}{gbsn}
\begin{tabular}{lp{0.34\linewidth}p{0.25\linewidth}p{0.27\linewidth}}
\toprule
 & \textbf{Sentence} & \textbf{Native gloss} & \textbf{LLM keywords} \\
\midrule
EN & Why? & --- & why \\
EN & I'm having fun. & --- & i fun \\
EN & I went there for a football game once. & --- & i go there football game once \\
EN & You don't have to take big steps. & --- & you not take big step \\
EN & And that's how you tune a ukulele. & --- & how you tune ukulele \\
EN & Hold down the second string on the fourth fret and then pluck the second and third string. & --- & hold down second string fourth fret pluck second third string \\
\midrule
DE & teilweise kr\"aftiger regen zieht heute nacht \"uber uns ostw\"arts & TEILWEISE REGEN KOMMEN IX & teilweise kr\"aftig regen ziehen heute nacht \"uber uns ostw\"arts \\
DE & in den \"ubrigen landesteilen wird es meist freundlich dank einer hochdruckzone die von der biskaya bis zu den shetlandinseln reicht & REGION MEHR FREUNDLICH HOCH DRUCK KOMMEN BIS REGION & \"ubrige landeteil werden freundlich dank hochdruckzone reichen biskaya shetlandinsel \\
\midrule
ZH & 我要去超市买椅子，你去吗？ & 超市 我 要 买 椅子 你 去 & 我 去 超市 买 椅子 你 去 \\
ZH & 性格，既不坚固也不是一成不变，而是活动变化着的。 & 性格 不是 稳定 不是 点 变化 没有 是 活动 变化 & 性格 不 坚固 也 不 一成不变 活动 变化 \\
\bottomrule
\end{tabular}
\end{CJK}
\end{table*}

\paragraph{Representative outputs.}
\Cref{tab:supp_kw_examples} shows uncurated examples in the three
languages next to the corpus's native gloss where one exists. The
extractor is
imperfect: occasional artefacts include over-eager merging of adjacent
tokens and sporadic lemmatisation slips. These are systematic but rare
and do not change the conditioning regime: on How2Sign no keyword
variant escapes collapse, and on Phoenix-2014T and CSL-Daily the
keyword variants stay in the same band as the sentence and native-gloss
variants (Table~3).

\paragraph{Fused input.}
The \emph{fused} variant concatenates the sentence and its keywords
(Phoenix-2014T) or native gloss (CSL-Daily) into one string,
``\texttt{sentence ; keywords}'', which is then encoded by the same
frozen text encoder as a single pooled token. Nothing else changes.

\paragraph{English prompt.}
The prompt below is sent to the LLM once per sentence, with
\texttt{\{input\_sentence\}} replaced by the source text and the
model's output read as the keyword string. Arrows in the original
prompt files are rendered as \texttt{->} here.

{\scriptsize\ttfamily\raggedright\obeylines\setlength{\parindent}{0pt}
You are an ASL (American Sign Language) linguistic expert. Your task is to extract pseudo-gloss from an English sentence.
\mbox{}
Pseudo-gloss is a sequence of words that determine the core meaning of the sentence. Each word should correspond to a distinct sign that an ASL signer would produce. The test is: ''If I remove this word, does the core meaning of the sentence change?''
\mbox{}
Follow these three steps in order:
\mbox{}
\#\# Step 1: Phrase-level chunking
Identify multi-word structures and treat them as a single unit BEFORE analyzing individual words:
- Auxiliary verb chains: ''is going to'', ''has been doing'', ''would have been'', ''are going to be learning'' -> mark the entire chain as auxiliary, drop it all
- Phrasal verbs: ''get out'', ''pick up'', ''break down'', ''set up'', ''hold down'' -> mark as a single verb unit, keep the particle
- Light verb constructions: ''make sure''->''sure'', ''take a look''->''look'', ''get rid of''->''rid'' — the light verb (make/take/get) is structural, drop it
- Discourse frames: ''let me/let's'' (when meaning ''allow me to''), ''you know'', ''I mean'' -> drop
\mbox{}
This step is critical: without it, ''going'' inside ''is going to'' would be mistakenly kept as ''go''.
\mbox{}
\#\# Step 2: Decide which units determine the core meaning
Keep words that would be produced as an independent sign in ASL:
- Content words: nouns, verbs, adjectives, content adverbs (always, sometimes, often)
- Pronouns: i, you, we, he, she, they, me, my, your, his, her, our, their
- Negation: not, no, never, nothing, none, nobody
- Interrogative WH-words: what, when, where, who, why, how, which (in questions)
- Numbers: one, two, three, second, third, etc.
- Modal verbs that carry meaning: can (ability), might/maybe (possibility), should/must (obligation)
- Spatial/directional words when they express location or movement: inside, outside, up, down, around, forward, back, off, over
\mbox{}
Drop words that don't change the core meaning:
- Articles: a, an, the
- Copulas: is, am, are, was, were (as linking verb)
- Auxiliary chains identified in Step 1
- Conjunctions: and, but, or, so, because, if, that, then
- Structural prepositions: of, to (infinitive marker), at, by, for, with, from (when purely grammatical)
- Degree modifiers: very, really, pretty (as ''fairly''), quite, extremely, a lot — these are expressed through sign intensity in ASL, not as separate signs
- Discourse fillers: okay, well, actually, basically, just, literally, essentially
- Determiners: this, that, these, those, any, some (when not carrying meaning)
- Relative pronouns: which/that introducing a clause (''the button, which opens...'' -> drop ''which'')
- Structural ''how'': ''learn how to play'' -> drop ''how'' (vs interrogative ''How do you play?'' -> keep ''how'')
\mbox{}
If the same content repeats for emphasis in the original sentence, keep it only once (de-duplicate).
\mbox{}
\#\# Step 3: Restore to citation form
Lemmatize each kept word to its base/dictionary form:
- Verb tense: gives->give, went->go, tied->tie, learning->learn, spoken->speak
- Plural: steps->step, hands->hand, feet->foot, children->child
- Comparative/superlative: highest->high, bigger->big, better->good
- Never produce a non-word fragment (e.g. other->oth is WRONG, keep ''other'')
\mbox{}
\#\# Output format
Output ONLY the pseudo-gloss as a single line of space-separated lowercase words. No explanation.
\mbox{}
\#\# Examples
\mbox{}
Sentence: I'm having fun.
Gloss: i fun
\mbox{}
Sentence: You don't have to take big steps.
Gloss: you not take big step
\mbox{}
Sentence: Hold down the second string on the fourth fret and then pluck the second and third string.
Gloss: hold down second string fourth fret pluck second third string
\mbox{}
Sentence: Today we're going to be learning how to play Portal, a game by Valve Software.
Gloss: today we learn play portal game valve software
\mbox{}
\textrm{\emph{[26 of 30 few-shot examples omitted for space]}}
\mbox{}
Sentence: \{input\_sentence\}
Gloss:
}

\paragraph{German prompt.}

{\scriptsize\ttfamily\raggedright\obeylines\setlength{\parindent}{0pt}
You are a DGS (German Sign Language / Deutsche Gebärdensprache) linguistic expert. Your task is to extract pseudo-gloss from a German sentence.
\mbox{}
Pseudo-gloss is a sequence of words that determine the core meaning of the sentence. Each word should correspond to a distinct sign that a DGS signer would produce. The test is: ''If I remove this word, does the core meaning of the sentence change?''
\mbox{}
Follow these three steps in order:
\mbox{}
\#\# Step 1: Phrase-level chunking
Identify multi-word structures and treat them as a single unit BEFORE analyzing individual words:
- Auxiliary chains: ''ist gegangen'', ''hat gemacht'', ''wird kommen'', ''war gewesen'', ''sind worden'' -> mark the entire chain as auxiliary, drop the auxiliary (sein/haben/werden), keep only the participle/infinitive lemma
- Modal + infinitive: ''kann gehen'', ''muss machen'', ''soll bleiben'' -> keep the modal only if it carries meaning (kann=ability, muss=obligation), keep the main verb
- Reflexive constructions: ''sich freuen'', ''sich ärgern'' -> drop ''sich'' (reflexive marker), keep the verb
- Particle verbs: ''aufstehen'', ''ankommen'', ''wegfahren'' -> keep as one unit (already a single lemma)
- Light verb constructions: ''Bescheid geben''->''geben/inform'', ''in Verbindung treten''->''verbinden'' — collapse to a single content word
- Discourse frames: ''und nun'', ''weiß ich nicht'', ''denke ich'' -> drop the frame, keep the proposition
\mbox{}
\#\# Step 2: Decide which units determine the core meaning
Keep words that would be produced as an independent sign in DGS:
- Content words: nouns (Wetter, Regen), verbs (gehen, regnen, sein-as-existence), adjectives (warm, kalt), content adverbs (heute, morgen, oft)
- Pronouns: ich, du, er, sie, wir, ihr, mein, dein, sein, unser — DGS expresses these via IX (indexing) and POSS signs
- Negations: nicht, nie, niemals, kein, keine, nichts, niemand, weder, ohne
- Interrogative WH-words: was, wann, wo, wer, warum, wie, welcher
- Numbers and quantifiers: ein/eine (when meaning ''one'', not as article), zwei, drei, viele, wenige, alle
- Modal verbs that carry meaning: kann (ability), muss/soll (obligation), darf (permission), will (desire)
- Spatial/directional words: oben, unten, hier, dort, vor, hinter, rechts, links — when expressing location
- Time expressions: heute, morgen, gestern, jetzt, bald, früh, spät
\mbox{}
Drop words that don't change the core meaning:
- Articles: der, die, das, den, dem, des, ein, eine, einen, einer, eines (when used as articles, not numerals)
- Copulas / structural sein: ist, sind, war, waren, sei, gewesen — when linking (''ist warm'' -> just ''warm'')
- Auxiliary chains identified in Step 1: haben + Partizip, werden + Infinitiv (future)
- Conjunctions: und, oder, aber, denn, weil, dass, obwohl, wenn, falls, sondern
- Structural prepositions: in, auf, an, von, mit, für, bei, zu, nach (when grammatical/structural)
- Degree modifiers: sehr, ziemlich, ganz, recht, durchaus — DGS expresses degree via sign intensity, not separate signs
- Discourse fillers: also, eigentlich, halt, eben, mal, doch, ja, schon, nun
- Vague quantifiers / particles: etwas, irgendwie, sozusagen
- Impersonal ''es'': ''es regnet'' -> just ''regnen''; ''es gibt'' -> ''geben/exist''
- Filler relative pronouns: der/die/das as relative pronouns introducing subordinate clauses
\mbox{}
If the same content repeats for emphasis in the original sentence, keep it only once (de-duplicate).
\mbox{}
\#\# Step 3: Restore to citation form (German lemmatization)
Lemmatize each kept word to its base/dictionary form:
- Verb conjugation: gehe/gehst/geht/gingen->gehen; regnet/regnete->regnen; bin/bist/ist/war->sein
- Plural: Tagen->Tag; Wolken->Wolke; Häuser->Haus; Männer->Mann
- Adjective inflection: tiefer/tiefen/tieferes->tief; kräftige/kräftigen->kräftig; nächsten->nächst
- Comparative/superlative: höher->hoch; größer->groß; am besten->gut
- Decapitalize all words. DGS gloss is conventionally upper-case, but for our pipeline we output lowercase for tokenizer compatibility.
\mbox{}
\#\# Output format
Output ONLY the pseudo-gloss as a single line of space-separated lowercase words. No explanation. No quotation marks. No leading punctuation.
\mbox{}
\#\# Examples
\mbox{}
Sentence: ich freue mich.
Gloss: ich freuen
\mbox{}
Sentence: und nun die wettervorhersage für morgen donnerstag den zwölften august
Gloss: wettervorhersage morgen donnerstag zwölft august
\mbox{}
Sentence: im nordwesten bleibt es heute nacht meist trocken sonst muss mit teilweise kräftigen schauern gerechnet werden
Gloss: nordwesten bleiben heute nacht trocken muss kräftig schauer rechnen
\mbox{}
Sentence: was ist mit dem wetter morgen?
Gloss: was wetter morgen
\mbox{}
\textrm{\emph{[3 of 7 few-shot examples omitted for space]}}
\mbox{}
Now extract the pseudo-gloss for the following German sentence.
\mbox{}
Sentence: \{input\_sentence\}
Gloss:
}

\paragraph{Chinese prompt.}
Note that the few-shot sentences of this prompt are deliberately
\emph{not} used as examples anywhere in the paper.

\begin{CJK}{UTF8}{gbsn}
{\scriptsize\ttfamily\raggedright\obeylines\setlength{\parindent}{0pt}
You are a CSL (Chinese Sign Language / 中国手语) linguistic expert. Your task is to extract keywords (pseudo-gloss) from a Chinese sentence.
\mbox{}
Keywords are a sequence of words that determine the core meaning of the sentence. Each word should correspond to a distinct sign that a CSL signer would produce. The test is: ''If I remove this word, does the core meaning of the sentence change?'' Keep the words in the ORIGINAL sentence order.
\mbox{}
Follow these three steps in order:
\mbox{}
\#\# Step 1: Phrase-level chunking
Identify multi-word structures and treat them as a single unit BEFORE analyzing individual characters:
- Fixed expressions: 对不起, 没关系, 不客气, 谢谢, 再见 -> keep as ONE unit, do not split
- Light verb constructions: ''打手语''->''手语'' (打 is structural), ''叫什么名字''->''什么 名字'' (叫 is structural), ''取名字''->keep the action verb 取 and 名字
- Aspect/structural particles attached to verbs: 了, 着, 过 -> drop the particle, keep the verb
- ''是……的'' emphasis frame -> drop 是 and 的, keep the content inside
- 把/被 constructions -> drop the marker 把/被, keep agent, patient, and verb
\mbox{}
\#\# Step 2: Decide which units determine the core meaning
Keep words that would be produced as an independent sign in CSL:
- Content words: nouns (房间, 名字, 工作), verbs (去, 认识, 介绍, 起床), adjectives (好, 小, 高兴)
- Pronouns: 我, 你, 他, 她, 我们, 你们, 他们, 这, 那 (when referring to a thing/place)
- Negation: 不, 没, 没有, 别, 从不
- Question words: 什么, 谁, 哪, 哪里, 几, 多少, 为什么, 怎么
- Numbers: 一, 二, 三, 十, 五十, 100 — keep the numeral, DROP the measure word after it (个, 只, 条, 本, 张, 位, 岁 keep 岁 only when age is the point of the sentence)
- Modal verbs that carry meaning: 会 (ability), 能/可以 (possibility/permission), 要/想 (desire), 应该/必须 (obligation)
- Time words: 今天, 明天, 昨天, 今年, 现在, 每天, 早上, 晚上
- Location/direction words when they express real location: 上, 下, 里, 外, 前, 后, 旁边, 家
\mbox{}
Drop words that don't change the core meaning:
- Structural particles: 的, 地, 得
- Aspect/sentence-final particles: 了, 着, 过, 吗, 呢, 吧, 啊, 呀
- Copula 是 when it just links (''他是老师'' -> 他 老师)
- Existence verb 有 when a number already carries the meaning (''我家有三个人'' -> 我 家 三 人)
- Measure words / classifiers: 个, 只, 条, 本, 张, 位, 件, 辆
- Conjunctions: 和, 跟, 或者, 但是, 因为, 所以, 如果, 虽然, 然后
- Structural prepositions: 在, 从, 到, 对, 给, 向, 跟 (when purely grammatical; keep when they express real location or destination)
- Degree adverbs: 很, 非常, 太, 特别, 真, 挺 — CSL expresses degree through sign intensity, not separate signs
- Discourse fillers and adverbs of emphasis: 就, 都, 也, 还, 又, 再, 其实, 当然
\mbox{}
If the same content repeats for emphasis, keep it only once (de-duplicate).
\mbox{}
\#\# Step 3: Output as segmented words
Chinese has no inflection, so no lemmatization is needed. Segment the kept content into standard Chinese words (not single characters unless the word is one character). Keep the original sentence order.
\mbox{}
\#\# Output format
Output ONLY the keywords as a single line of space-separated Chinese words. No explanation. No quotation marks. No punctuation.
\mbox{}
\#\# Examples
\mbox{}
Sentence: 你叫什么名字？
Keywords: 你 什么 名字
\mbox{}
Sentence: 我家有三个人。
Keywords: 我 家 三 人
\mbox{}
Sentence: 你会打手语吗？
Keywords: 你 会 手语
\mbox{}
Sentence: 这是我和我丈夫的房间，旁边那个小的房间是我女儿的。
Keywords: 这 我 丈夫 房间 旁边 小 房间 我 女儿
\mbox{}
\textrm{\emph{[12 of 16 few-shot examples omitted for space]}}
\mbox{}
Now extract the keywords for the following Chinese sentence.
\mbox{}
Sentence: \{input\_sentence\}
Keywords:
}
\end{CJK}

\section{Back-translation evaluator: training details and failure modes}
\label{sec:supp_bt}

\paragraph{Training setup.}
Our back-translation (motion-to-text) evaluator on How2Sign is
trained pose-to-text with translation cross-entropy only, no
gloss-CTC loss, no pretrained CSLR features, no auxiliary
pretraining. Input is axis-angle SMPL-X; decoding at evaluation is
beam-$5$; the test split has $n = 2{,}357$ paired sentences. In
this regime BT-BLEU-4 plateaus at a ceiling of $\approx 2.5$ even on
GT motion. Without pretraining, How2Sign alone does not train a usable
translation network, and cross-regime comparisons against
gloss-CTC-supervised evaluators (e.g.\ Sign Language
Transformers~\cite{camgoz2020slt}) are not directly meaningful.

\paragraph{Failure-mode commentary.}
Table~4 shows $3$ representative GT/prediction
pairs for our generated motion (BLEU-4 $= 1.24$). The no-pretraining
video translator of~\citet{youtubeasl2023} reaches a similar
BLEU-4 $\approx 1.22$. The pattern is consistent across the test
split: (i) input-independent high-frequency template language ---
instructional constructions (``you can do it'', ``you want to make
sure'', ``i'm going to take my right/left hand'') and pronoun-heavy
filler --- whose semantic content is unrelated to the GT motion; and
(ii) degenerate end-of-sequence repetition (``very light, very light,
very light, $\ldots$''). Both patterns are characteristic of an
autoregressive decoder estimating a near-marginal language-model prior
with negligible conditional signal from the encoded motion. A non-zero
BLEU score arises from short n-gram overlap with How2Sign's
high-frequency vocabulary, not from recovered text content; the
diagnostic counterpart, $\rfaith$, captures this failure directly.

\section{Implementation details}
\label{sec:supp_impl}

\paragraph{MoAE.}
The MoAE $\phi$ of the diagnostic is a small transformer autoencoder
operating on per-frame $3$-D joint coordinates: each SMPL-X axis-angle
sequence is first lifted to $T \times 144 \times 3$ joints by SMPL-X
forward kinematics, restricted to the $44$ upper-body joints (spine,
head, shoulders, elbows, wrists, all $30$ finger joints), and made
pelvis-relative, yielding per-frame input dimension $D = 132$. A
linear projection lifts each frame to a $d_{\mathrm{model}} = 256$
token; learnable absolute positional embeddings are added; a $3$-layer
transformer encoder ($4$ attention heads, feed-forward dim $512$,
GELU, pre-LN) produces contextualized per-frame features. A masked
mean over valid frames (padding excluded) gives the $256$-dim
bottleneck $z$. The decoder broadcasts $z$ across the target
sequence length, adds the same positional embeddings, runs through
another $3$-layer transformer stack with the same hyper-parameters,
and projects back to $D$. Position embeddings are shared between
encoder and decoder; padding is handled with a boolean mask in both
directions.

\textbf{Training loss.} For each batch we minimize per-frame MSE
between reconstruction and target plus an
$\alpha_{\mathrm{vel}} = 0.5$ MSE on the first-difference (velocity)
of the reconstruction:
\begin{equation*}
\mathcal{L}_{\mathrm{MoAE}}
\;=\; \mathrm{MSE}(\hat{x}, x)
\;+\; \alpha_{\mathrm{vel}} \, \mathrm{MSE}(\Delta \hat{x}, \Delta x).
\end{equation*}
The velocity term is essential: without it, the MoAE collapses to
emitting a per-sample mean pose at every frame, which would make $z$
insensitive to motion dynamics and defeat the diagnostic; the
encoder must retain information about \emph{how} the body moves, not
only \emph{what} poses it visits. Training: the target dataset's
train motions only, sequence length $T = 200$ frames, batch size
$128$, $100$ epochs, AdamW~\cite{loshchilov2019adamw} at lr
$3\times 10^{-4}$, dropout $0.1$ (eval at dropout $0$), bf16 mixed
precision. The encoder is never exposed to model-generated motion
during training, so the bottleneck does not learn collapse-specific
features that could bias the diagnostic toward any particular
generator. The diagnostic has the limitations of any MoAE: a
poorly-trained MoAE could in principle push $\rfaith$ artificially
close to $1$. We address this with a single frozen MoAE trained on the
target dataset's GT train motions, the MoAE-initialization and
GT-vs-GT checks of Sec.~3.2, and an independent check
against visual classification of rendered outputs, which gives the
same conclusion on every checkpoint where we did it. One MoAE of this
architecture and recipe is trained per
dataset (How2Sign, Phoenix-2014T, CSL-Daily, ASL3DWord), and the
\emph{same} frozen checkpoint scores every diagnostic on that dataset,
removing per-paper feature inconsistency. A further MoAE of the same
architecture, trained on Phoenix-2014T MediaPipe keypoints
($7{,}096$ sequences), is used only for the PT experiment of
\cref{sec:supp_pt}.

\paragraph{Cross-dataset ratios are qualitative, not quantitative.}
The ratios $\rdiv$, $\rfaith$, $\rcond$ are dimensionless within one
$\phi$: numerator and denominator are L\textsubscript{2} distances in
the \emph{same} latent space, so the random-pairing baseline cancels
the absolute scale. Therefore, within a single dataset, an $\rfaith$
of $0.73$ vs $0.96$ means the model is genuinely more faithful, not
that one $\phi$ was ``stricter''. Across datasets, however, the latent
geometries differ and the same numerical $\rfaith$ value need not
correspond to identical motion quality. We use cross-dataset
comparisons \emph{qualitatively} only, to argue band membership
relative to each dataset's own GT-vs-GT reference
(\cref{sec:supp_ref}): ASL3DWord $0.73$ against its reference of
$0.752$; Phoenix-2014T $0.81$--$0.85$ against $0.874$; CSL-Daily
$0.75$--$0.80$ against $0.673$; How2Sign $0.89$--$1.00$ against
$0.849$ --- never as an ordinal ranking of two distinct-dataset
checkpoints by raw $\rfaith$.

\paragraph{ASL3DWord MoAE calibration.}
The diffusion checkpoint's $\rfaith = 0.73$ on gloss-level ASL3DWord
(row 16 of \cref{tab:supp_main}) is interpreted against the
within-ASL3DWord GT-vs-GT reference: $339$ test motions are encoded by
the same ASL3DWord MoAE used elsewhere; pairwise L\textsubscript{2}
distances in the latent are partitioned into \emph{same-gloss} pairs
($874$ pairs, both encodings GT clips of the same target gloss) and
\emph{different-gloss} pairs ($113{,}708$). The same-gloss mean
distance ($0.826$) divided by the different-gloss mean distance
($1.098$) gives $\rfaithref = 0.752$, the lowest $\rfaith$ achievable
on this MoAE under perfect within-class generation. The checkpoint's
$0.73$ is $0.022$ below this value, meaning generated motion is, in the
MoAE latent, on average as close to its target as a second GT clip of
the same gloss. Consequently the gap between ASL3DWord ($0.73$) and the
How2Sign band ($[0.89, 1.00]$) \emph{understates} the actual difference:
How2Sign checkpoints sit $0.04$--$0.15$ above their $0.849$ reference,
ASL3DWord is at its reference.

\paragraph{Data preparation for Phoenix-2014T and CSL-Daily.}
Both corpora use the per-frame SMPL-X fittings released with Signs as
Tokens~\cite{zuo2025soke}, aggregated per video into axis-angle
sequences with the same layout as How2Sign ($53$ joints
$\times\,3$). Global orientation is zeroed for both GT and generated
motions before encoding (the generators are root-normalized). Sequences
are resampled to $T = 100$ frames by uniform index sampling. The same
$3$-to-$30$ length filter as How2Sign is applied to the
\emph{original} sentence (words for German, characters for Chinese),
never to the keyword or gloss string, giving $7{,}003$ train / $641$
test sentences on Phoenix-2014T and $18{,}399$ train / $1{,}176$ test
on CSL-Daily; every conditioning variant of a corpus therefore trains
and is evaluated on the identical sample set. Phoenix sentences are the
German \texttt{translation} field and native gloss is the \texttt{orth}
field of the corpus annotation ($\sim$$1{,}066$ DGS gloss types over
$\sim$$7$K sentence--motion pairs, against $\sim$$15$K English word
types over $\sim$$31$K pairs on How2Sign); CSL-Daily sentences and
gloss are the
dataset's own annotations.

\begin{figure*}[t]
\centering
\begin{tikzpicture}[
  font=\scriptsize, x=1mm, y=1mm,
  blk/.style={draw, rounded corners=1.5pt, inner sep=2.5pt, align=center, minimum height=9mm, minimum width=22mm},
  io/.style={blk, fill=gray!12},
  frz/.style={blk, fill=yellow!20},
  tr/.style={blk, fill=blue!10, minimum width=36mm, minimum height=11mm},
  zb/.style={blk, fill=green!15},
  arr/.style={->, >=Latex, thin},
  lab/.style={font=\scriptsize\bfseries, anchor=west}
]
\node[lab] at (0,30) {(a) MDM denoiser (diffusion)};
\node[io]  (a1) at (12,22) {noisy motion $x_t$\\($T{\times}D$)};
\node[blk] (a2) at (40,22) {linear $D{\to}512$\\$+$ pos.\ emb.};
\node[tr]  (a3) at (76,22) {Transformer encoder\\$4$ layers, $d{=}512$, $8$ heads\\input $[t;\,c;\,x^{(1)}\!\ldots x^{(T)}]$};
\node[blk] (a4) at (112,22) {linear $512{\to}D$};
\node[io]  (a5) at (140,22) {$\hat{\epsilon}$ ($T{\times}D$)};
\draw[arr] (a1)--(a2); \draw[arr] (a2)--(a3); \draw[arr] (a3)--(a4); \draw[arr] (a4)--(a5);
\node[io]  (a6) at (12,8)  {text};
\node[frz] (a7) at (40,8)  {frozen text encoder\\(CLIP / Chinese-CLIP)};
\node[blk, minimum width=18mm] (a8) at (67,8)  {token $c$\\(dropped w.p.\ $0.1$)};
\node[blk, minimum width=18mm] (a9) at (87,8)  {timestep\\token $t$};
\draw[arr] (a6)--(a7); \draw[arr] (a7)--(a8);
\draw[arr] (a8.north) -- (a8.north |- a3.south);
\draw[arr] (a9.north) -- (a9.north |- a3.south);
\node[lab] at (0,-8) {(b) Clean transformer-decoder regression};
\node[io]  (b1) at (12,-16) {$T$ learned\\frame queries};
\node[tr]  (b2) at (54,-16) {Transformer decoder\\$6$ layers, $d{=}512$, $8$ heads\\self-attn $+$ cross-attn to $c$};
\node[blk] (b3) at (90,-16) {linear $512{\to}D$};
\node[io]  (b4) at (118,-16) {motion ($T{\times}D$)};
\draw[arr] (b1)--(b2); \draw[arr] (b2)--(b3); \draw[arr] (b3)--(b4);
\node[io]  (b5) at (12,-30) {text};
\node[frz] (b6) at (40,-30) {frozen CLIP};
\node[blk, minimum width=18mm] (b7) at (62,-30) {token $c$\\(memory)};
\draw[arr] (b5)--(b6); \draw[arr] (b6)--(b7);
\draw[arr] (b7.north) -- (b7.north |- b2.south);
\node[lab] at (0,-46) {(c) MoAE (frozen, one per dataset; used only by the diagnostic)};
\node[io]  (c1) at (12,-54) {SMPL-X motion\\($T{\times}159$)};
\node[blk] (c2) at (40,-54) {forward kinematics\\$44$ upper-body joints\\pelvis-relative ($T{\times}132$)};
\node[blk] (c3) at (68,-54) {linear $132{\to}256$\\$+$ pos.\ emb.};
\node[tr]  (c4) at (102,-54) {Transformer encoder\\$3$ layers, $d{=}256$, $4$ heads};
\node[zb]  (c5) at (136,-54) {masked mean\\$z\in\mathbb{R}^{256}$};
\draw[arr] (c1)--(c2); \draw[arr] (c2)--(c3); \draw[arr] (c3)--(c4); \draw[arr] (c4)--(c5);
\node[blk] (c6) at (136,-68) {broadcast $z$ over $T$\\$+$ pos.\ emb.};
\node[tr]  (c7) at (102,-68) {Transformer decoder\\$3$ layers, $d{=}256$, $4$ heads};
\node[blk] (c8) at (68,-68) {linear $256{\to}132$};
\node[io]  (c9) at (40,-68) {reconstruction\\($T{\times}132$)};
\draw[arr] (c5)--(c6); \draw[arr] (c6)--(c7); \draw[arr] (c7)--(c8); \draw[arr] (c8)--(c9);
\end{tikzpicture}
\caption{Architectures used in the paper. (a)~MDM denoiser: the
frozen text encoder yields one conditioning token $c$ (CLIP-ViT-B/32
for English/German, Chinese-CLIP for Chinese; dropped with
probability $0.1$ for classifier-free guidance); $c$ and a timestep
token are prepended to the $T$ projected motion tokens, a $4$-layer
Transformer encoder ($d{=}512$, $8$ heads, dropout $0.1$) processes
the sequence, and the motion positions are projected back to
$\hat{\epsilon}$ ($\epsilon$-prediction; cosine schedule, $1{,}000$
training steps, $50$-step DDIM sampling with guidance $3.0$).
(b)~Clean regression: $T$ learned frame queries pass through a
$6$-layer Transformer decoder whose cross-attention memory is the
single token $c$; one forward pass emits the full sequence
($\sim$$26$M trainable parameters). (c)~MoAE: motions are lifted by
forward kinematics to $44$ pelvis-relative upper-body joints, encoded
by a $3$-layer Transformer, mean-pooled over valid frames into
$z \in \mathbb{R}^{256}$, and decoded by a symmetric $3$-layer stack;
the diagnostic uses only $z$.}
\label{fig:supp_arch}
\end{figure*}

\paragraph{Diffusion variants.}
\Cref{fig:supp_arch} draws the three networks used throughout the
paper: the MDM denoiser (a), the clean regression model (b), and the
MoAE (c). The diffusion experiments cover $2$ backbones and $2$--$4$
conditioning strings (plus a phonology-augmented variant), all on a
standard MDM~\cite{tevet2023mdm} base (\cref{fig:supp_arch}a: $4$-layer
Transformer encoder, $d{=}512$, $8$ heads, timestep and condition
tokens prepended, $\epsilon$-prediction). For every component below the finding
on How2Sign is the same: \emph{$\rfaith$ remains in the marginal-collapse
band $[0.89, 1.00]$ regardless of which combination is used.}

\textbf{Backbones (2).}
\emph{Standard}: the MDM denoiser of \cref{fig:supp_arch}a.
\emph{Multi-scale temporal transformer (\textsf{MSTrans})}: the
same denoiser with the encoder replaced by a $4$-scale temporal encoder at frame-rate factors $1, 2, 4, 8$ with
cross-scale fusion, intended to capture multi-rate motion structure
(slow trajectory $+$ fast hand articulation): the noisy motion is
processed at strides $1, 2, 4, 8$ by four independent Transformer
encoders, each cross-attending to the text token, and the per-scale
features are upsampled to $T$ frames and fused. It changes FID by
$<0.2$ and $\rfaith$ by $\le 0.02$ (\cref{tab:supp_main}, rows 3--4).

\textbf{Conditioning inputs.} \emph{Raw sentence}, encoded
by the frozen text encoder and pooled to a single token; \emph{LLM
keywords} (\cref{sec:supp_kw}); on the gloss-annotated corpora also
\emph{native gloss} and the \emph{fused} string. The text encoder is
CLIP-ViT-B/32~\cite{radford2021clip} for English and German and the
Chinese-CLIP text tower~\cite{yang2022chineseclip} for Chinese, using
its CLS token after the contrastive projection (mean-pooled hidden
states form a new anisotropic cone and are not used). On How2Sign we
additionally tested a \emph{phonology-augmented} keyword input with
per-token phonological-attribute embeddings
following~\citet{hong2026signphonology}; its collapse pattern is
unchanged and it is not separately tabulated.

\textbf{Loss and sampling.} Per-frame pose MSE plus an FK
joint-position MSE term. All MDM variants are trained with
classifier-free guidance ($10\%$ condition dropout) and sampled with
guidance scale $3.0$; the Phoenix-2014T and CSL-Daily runs use
$T = 100$, batch size $64$, and $100$ epochs, identical across the
sentence, keyword, gloss, and fused variants of a corpus.

\paragraph{Clean regression model.}
The clean regression model is a $6$-layer Transformer
decoder~\cite{vaswani2017attention} with $D = 512$, $8$ heads, $0.1$
dropout, GELU. The decoder's queries are $T$ learnable positional
embeddings; cross-attention memory is a single text token from the
frozen text encoder. The output projects to $132$
upper-body axis-angle dimensions. Training: AdamW
($\beta = (0.9, 0.999)$, weight decay $10^{-2}$),
\texttt{OneCycleLR} schedule with $5\%$ warmup and cosine annealing,
fp16 mixed precision. Loss is
$\mathcal{L}_{\mathrm{pose}} + \lambda_{\mathrm{FK}}\,\mathcal{L}_{\mathrm{FK}}$
(per-frame pose MSE and FK joint-position MSE) with
$\lambda_{\mathrm{FK}} = 5$. Total trainable parameters $\sim$$26$M
(with frozen text encoder excluded). Phoenix-2014T and CSL-Daily
regression runs use batch size $64$, lr $10^{-4}$, and $100$ /
$65$ epochs respectively (the CSL-Daily evaluation loss plateaus from
epoch $\sim$$70$; the reported checkpoint is the best-validation one).

\paragraph{Differences from Progressive Transformers (PT).}
The clean regression model differs from
PT~\cite{saunders2020progressive} in four ways:
(i)~we generate all $T$ frames in a single forward pass, whereas PT
is autoregressive (one frame at a time);
(ii)~output length $T$ is fixed per sample, whereas PT uses a
learned step counter to decide when to stop, which the released
protocol drives with GT timing at inference (\cref{sec:supp_pt});
(iii)~the text encoder is frozen CLIP (a single pooled
token), whereas PT trains its own text encoder over T5-style
tokenization;
(iv)~the decoder sees only the text embedding via cross-attention,
with no feedback from previously generated frames.
These simplifications remove any dependence on GT timing at
inference and let the row report what a basic
\emph{Transformer-decoder $+$ text cross-attention} recipe achieves
at the same paired-data budget.

\paragraph{Vocabulary-restricted training subsets.}
The $K{=}3000$ and $K{=}500$ whitelists, as well as the $\le 1$-word
and $\le 3$-word short-sentence whitelists, are derived from
frequencies on How2Sign train. The $K{=}K$ filter keeps only
sentences whose words are all within the top-$K$ most-frequent
training-vocabulary words. \emph{The whitelist is always defined on
the raw-sentence vocabulary} regardless of whether the model receives
the sentence or the LLM keywords as input; the \texttt{sent} and
\texttt{keywords} variants in \cref{tab:supp_main} therefore share the
same set of training and test sentences and differ only in the
conditioning modality. For each variant we store the train and test
sentence lists under the same train-frequency vocab, ensuring
within-distribution evaluation. Counts (train / test): $K{=}3000$:
$13{,}571 / 890$; $K{=}500$: $2{,}918 / 241$; $\le 3$-word:
$1{,}502 / 121$; $\le 1$-word: $564 / 44$.

\section{Per-dataset GT-vs-GT reference protocol}
\label{sec:supp_ref}

Eq.~(9) defines $\rfaithref$ as the mean latent distance
between two GT renditions of the same text divided by the mean latent
distance between renditions of different texts. All references use the
dataset's own frozen MoAE, the same root-zeroing and $T = 100$
resampling as the diagnostic, and pairs drawn exactly as for the
generator (numerator and denominator on the same rendition pool).
Table~1 lists the group and pair counts; the mean
latent distances (same-text / different-text pairs) behind each value
are ASL3DWord $0.826/1.098$, Phoenix-2014T $2.167/2.479$, CSL-Daily
$1.236/1.836$, How2Sign $3.346/3.940$ (all cross-video repeats:
$5.230/5.248$).

\paragraph{ASL3DWord.} Groups are glosses; $100$ of the $103$ test
glosses have $\geq 2$ clips (different signers). Same-gloss pairs are
across signers by construction.

\paragraph{Phoenix-2014T.} Groups are exact German sentences recurring
with $\geq 2$ renditions in the pooled train$+$test split (dev
excluded); $50$ such sentences exist, mostly formulaic broadcast
openers and closers. Many renditions share a signer, which makes the
reference conservative (same-signer pairs are closer than cross-signer
pairs, lowering $\rfaithref$ and making it \emph{harder} to reach).

\paragraph{CSL-Daily.} Every training sentence is re-performed by
several signers ($2.8$ on average, up to $6$), so same-text groups are
abundant ($\sim$$6.5$K); we sample $1{,}000$ groups (seed fixed) from
the pooled train$+$test split, dev excluded. $99.6\%$ of same-sentence
pairs are cross-signer (only $9$ of $2{,}504$ share a signer): the CSL
reference measures scripted cross-signer consistency, the strictest of
the four.

\paragraph{How2Sign.} Pooled train$+$test contains $789$ exact-repeat
groups ($2{,}338$ renditions), but most are within-video: re-alignment
overlaps, duplicated metadata rows, or same-take repeats (e.g.\
exercise counting), which carry no independent rendition. Restricting
to repeats across \emph{distinct} videos leaves $181$ groups
($923$ renditions), dominated by one- and two-word greetings and
fillers (``Hi!'', ``Okay?'', ``Cheers!'') whose clips carry no
sentence-specific motion: same-text pairs are as far apart as random
pairs and the reference reads $0.997$, i.e.\ the metric correctly
reporting a signal-free condition. We therefore report the reference on
the $50$ sentences of $\geq 5$ words recurring across distinct videos
($116$ renditions; typically intro/outro scripts re-performed across a
signer's video series), which gives $0.849$, consistent with the
Phoenix and ASL3DWord values. Both numbers are disclosed; all $14$
How2Sign checkpoints ($0.89$--$1.00$) lie above the $0.849$
reference.

\section{Coverage analyses and controls}
\label{sec:supp_cov}

\paragraph{Conditioning-space dispersion.}
Two measures per corpus, computed on the same frozen text encoder the
generators condition on (English CLIP-ViT-B/32 unless noted, so the
numbers describe the geometry of the signal the denoiser receives;
the last row re-measures CSL-Daily on the Chinese-CLIP text tower
that the reported CSL-Daily models condition on):
(a)~\emph{within-train dispersion}, the
pairwise cosine among training-sentence embeddings (subsample of
$4{,}000$ sentences, $\sim$$8$M pairs); (b)~\emph{coverage}, each
test sentence's cosine to its nearest training neighbor (full pool).
Both are reported raw (as the model sees them) and mean-centered
(train centroid subtracted and renormalized, which removes the
encoder's anisotropy). XLM-R~\cite{conneau2020xlmr} mean-pooled
embeddings are degenerate in the raw space ($\sim$$0.99$ cosine for
every pair, the well-known cone effect) and are therefore reported
centered only, as a direction-preserving control. \Cref{tab:supp_disp}
gives the full numbers behind Table~5.

\begin{table*}[t]
\centering
\small
\setlength{\tabcolsep}{5pt}
\caption{Conditioning-space dispersion and coverage, full numbers.
Within: mean $\pm$ std pairwise cosine among training sentences (raw
CLIP). Coverage: test$\to$nearest-train cosine, raw mean (5th
percentile / median) and mean-centered mean; last column mean-centered
coverage under XLM-R. Train / test counts in parentheses.}
\label{tab:supp_disp}
\begin{tabular}{lcccc}
\toprule
Corpus & Within (raw) & Cov.\ raw (p5 / med) & Cov.\ cent. & XLM-R cent. \\
\midrule
How2Sign (30,912 / 2,314) & $0.449 \pm 0.129$ & 0.800 (0.64 / 0.80) & 0.625 & 0.713 \\
Phoenix-2014T (7,092 / 642) & $0.729 \pm 0.085$ & 0.947 (0.89 / 0.95) & 0.794 & 0.731 \\
CSL-Daily, sentence (18,399 / 1,176) & $0.882 \pm 0.052$ & 0.964 (0.93 / 0.97) & 0.663 & 0.624 \\
CSL-Daily, sentence (Chinese-CLIP) & $0.720 \pm 0.089$ & 0.920 (0.84 / 0.93) & 0.710 & --- \\
\bottomrule
\end{tabular}
\end{table*}

\paragraph{Nearest-neighbor retrieval sweep.}
For each test sentence we retrieve the nearest training sentence by
cosine in the text-encoder space, take that training sample's GT
motion as the ``prediction'', encode it with the dataset's MoAE, and
compute $\rfaith$ and $\rdiv$ exactly as for a generator. The pool is
a random subset of the training split of the stated size (full test
split throughout). \Cref{tab:supp_nnret} lists the curves. On How2Sign
the curve is flat at $0.97$--$1.00$ over a $50\times$ pool increase; a
per-sample check at the full pool shows the distribution of
$d(\text{retrieved},\text{target})$ relative to random pairing is
indistinguishable from the random-pairing null at every quantile, so
no subpopulation of test sentences finds a matching training motion.
On Phoenix-2014T the curve improves with pool size to the $0.874$
reference. On CSL-Daily, English CLIP gives a flat curve near $0.93$,
while Chinese-CLIP retrieves better neighbors at every pool size
($0.888 \to 0.868$, minimum $0.854$ at $9$K) --- the model-free
signature of the encoder contamination discussed in the main text.

\begin{table}[t]
\centering
\small
\setlength{\tabcolsep}{3pt}
\caption{Nearest-neighbor retrieval $\rfaith$ as a function of
training-pool size (pool size: $\rfaith$).}
\label{tab:supp_nnret}
\begin{tabular}{lp{0.54\linewidth}}
\toprule
Corpus (encoder) & Curve \\
\midrule
How2Sign (CLIP) & 500: 0.967; 1K: 0.999; 2K: 0.984; 5K: 0.981; 13K: 0.972; 26K: 0.981 \\
Phoenix-2014T (CLIP) & 500: 0.905; 1K: 0.912; 2K: 0.886; 4K: 0.862; 7K: 0.867 \\
CSL-Daily (English CLIP) & 1K: 0.941; 5K: 0.925; 9K: 0.928; 18K: 0.928 \\
CSL-Daily (Chinese-CLIP) & 1K: 0.888; 5K: 0.877; 9K: 0.854; 18K: 0.868 \\
\bottomrule
\end{tabular}
\end{table}

\paragraph{Text-shuffling control.}
For each test sentence $t_i$ we generate with the sentence-conditioned
MDM (guidance $3.0$; How2Sign: the unfiltered $31$K checkpoint, row 6
of \cref{tab:supp_main}; Phoenix-2014T: row 6 of Table~3)
under four conditions: (a)~$t_i$ with seed $s_1$;
(b)~$t_i$ with seed $s_2$; (c)~a random word permutation of $t_i$
with seed $s_1$; (d)~a different test sentence $t_j$ with seed $s_1$
(matched seed). In the MoAE latent, $d_{\rm intra}$ is the mean
distance (a)--(b), $d_{\rm shuf}$ the mean distance (a)--(c), and
$d_{\rm inter}$ the mean distance (a)--(d). Sentences with fewer than
three words cannot be meaningfully shuffled and are dropped ($31$ of a
$1{,}000$-sentence compute subsample on How2Sign; $1$ of $642$ on
Phoenix-2014T). $\rfaith$ is computed for the original and the shuffled
conditioning against the same GT. \Cref{tab:supp_shuffle} gives the
numbers behind Sec.~6.4: on How2Sign, replacing the sentence
moves the output no more than re-sampling the noise
($d_{\rm inter}/d_{\rm intra} = 1.05$), whereas on Phoenix-2014T it
moves it well beyond noise ($1.35$) and shuffling the words moves it
less than a new seed ($0.49$).

\begin{table*}[t]
\centering
\small
\setlength{\tabcolsep}{6pt}
\caption{Text-shuffling control (sentence-conditioned MDM, full test
split minus unshufflable sentences).}
\label{tab:supp_shuffle}
\begin{tabular}{lrrrrrrc}
\toprule
Corpus & $n$ & $d_{\rm intra}$ & $d_{\rm shuf}$ & $d_{\rm inter}$ &
$\frac{d_{\rm shuf}}{d_{\rm intra}}$ & $\frac{d_{\rm inter}}{d_{\rm intra}}$ &
$\rfaith$ orig $\to$ shuf \\
\midrule
How2Sign & 969 & 2.98 & 2.08 & 3.13 & 0.70 & 1.05 & 0.963 $\to$ 0.970 \\
Phoenix-2014T & 641 & 2.48 & 1.22 & 3.34 & 0.49 & 1.35 & 0.838 $\to$ 0.855 \\
\bottomrule
\end{tabular}
\end{table*}

\paragraph{Encoder-swap control.}
The $26$K How2Sign clean regression model was retrained with a frozen
T5-base encoder~\cite{raffel2020t5} in place of CLIP (all else equal;
batch size $512$, T5 forward pass in fp32). Full test split ($n = 2{,}314$ sentences with valid SMPL-X fits, the
set every How2Sign diagnostic uses): $\rfaith = 0.942$, $\rdiv = 0.32$, versus $0.954$ /
$0.34$ with CLIP; best evaluation loss $0.0920$ vs $0.0925$. The
stronger encoder reaches the same solution and the same collapse.

\paragraph{Chinese-CLIP for CSL-Daily.}
English CLIP on Chinese byte-pair tokens yields a within-train cosine of
$0.88$ (\cref{tab:supp_disp}), i.e.\ nearly constant conditioning. All
CSL-Daily generators and the retrieval sweep were therefore re-run with
the Chinese-CLIP text tower~\cite{yang2022chineseclip} (CLS token after
the contrastive projection; the projected space separates distinct
sentences at cosine $\approx 0.79$, whereas mean-pooled hidden states
form a new cone at $0.92$ and are not used). Every CSL-Daily model
improved by $0.05$--$0.12$ in $\rfaith$ under the swap (sentence
$0.867 \to 0.798$, keywords $0.870 \to 0.787$, native gloss
$0.836 \to 0.788$, fused $0.865 \to 0.750$, regression
$0.872 \to 0.775$); only the Chinese-CLIP numbers are reported in the
main text. The dispersion measurement itself, re-run on the
Chinese-CLIP tower (last row of \cref{tab:supp_disp}), tells the same
story: within-train cosine falls from $0.88$ to $0.72$ and centered
coverage rises to $0.710$, between How2Sign and Phoenix-2014T ---
consistent with CSL-Daily's intermediate model outcome.

\section{Frame-level DTW alignability}
\label{sec:supp_dtw}

\paragraph{Protocol.}
GT and generated sequences are forward-kinematics lifted to the same
$44$ upper-body $3$D joints the MoAE consumes (root zeroed, $T = 100$),
but compared \emph{per frame}: the frame distance is the L\textsubscript{2}
distance between the two $44 \times 3$ joint sets, and the
alignment cost is the mean frame distance along the optimal
dynamic-time-warping path. We report the ratio of this cost for the
generation and its own GT to the cost for the generation and the GT of
an unrelated sentence (\emph{DTW/random}; $1$ = chance). The
\emph{equal-timestamp} variant uses the diagonal path instead of the
DTW path (\emph{diag/random}). The human--human baseline applies the
identical statistic to two GT renditions of the same sentence ($40$
same-text groups per corpus, pooled train$+$test); the generation
statistic uses $40$ test sentences (seed $42$; unseen sentences for
CSL-Daily), a sample disjoint from the four rendered sentences per
corpus.

\begin{table}[t]
\centering
\scriptsize
\setlength{\tabcolsep}{2pt}
\caption{Frame-level alignability (lower = more alignable). DTW/random
statistics over $40$ pairs each; the last column is the equal-timestamp
(no DTW) mean.}
\label{tab:supp_dtw}
\begin{tabular}{llrrrrr}
\toprule
Corpus & Pair & mean & median & p10 & p90 & diag mean \\
\midrule
Phoenix-2014T & GT vs GT (human) & 0.743 & 0.756 & 0.520 & 0.893 & 0.918 \\
Phoenix-2014T & Gen vs GT & 0.859 & 0.863 & 0.743 & 0.988 & 0.983 \\
CSL-Daily & GT vs GT (human) & 0.585 & 0.578 & 0.444 & 0.702 & 0.782 \\
CSL-Daily & Gen vs GT & 0.756 & 0.753 & 0.617 & 0.896 & 0.966 \\
\bottomrule
\end{tabular}
\end{table}

\paragraph{Findings.}
\Cref{tab:supp_dtw} gives the statistics behind Sec.~6.2.
(i)~Two humans are far from frame-identical ($0.74$ on Phoenix-2014T,
$0.59$ on CSL-Daily), so a single GT video demands more consistency
than signers exhibit. (ii)~Generation sits one band above the human
distribution on both corpora; normalized to each corpus's own human
mean, Phoenix-2014T stands at $1.16\times$ and CSL-Daily at
$1.29\times$, the same ordering as the sequence-level MoAE verdicts.
(iii)~The best-aligned tail of the generation distribution (p10)
reaches the human mean, and it is populated by the densest conditions:
in the $40$-sentence Phoenix sample, the broadcast sign-off
\emph{``ich w\"unsche noch einen sch\"onen abend''} occupies two of the
three best-aligned slots ($0.68$, $0.70$), and among the four rendered
Phoenix sentences the recurring opener is the only one inside the
human band ($0.66$; the three non-formulaic ones sit at
$0.93$--$0.99$). (iv)~Without DTW, timing offsets masquerade as pose
error: the equal-timestamp ratio is $0.98$ / $0.97$ for generations and
inflates even the human--human pairs to $0.92$ / $0.78$. GT-timeline
frame strips therefore overstate mismatch, and all figures show
DTW-matched pairs.

\paragraph{Rendered examples.}
\Cref{fig:supp_matched} shows the two rendered sentences not included
in Fig.~3, at the generation median (CSL-Daily) and in the
typical non-formulaic Phoenix range. Of the four rendered sentences per
corpus, those shown are: Phoenix-2014T ---
\emph{``und nun die wettervorhersage f\"ur morgen donnerstag den dritten
dezember''} ($0.663$; main Fig.~3a) and
\emph{``teilweise kr\"aftiger regen zieht heute nacht \"uber uns
ostw\"arts''} ($0.975$; \cref{fig:supp_matched}a); CSL-Daily ---
\begin{CJK}{UTF8}{gbsn}
性格，既不坚固也不是一成不变，而是活动变化着的。
\end{CJK}
(``Character is neither fixed nor immutable, but active and changing.'';
$0.699$; main Fig.~3b) and
\begin{CJK}{UTF8}{gbsn}
合肥气温低，你骑自行车上学一定很冷。
\end{CJK}
(``Hefei is cold; cycling to school must be freezing.''; $0.743$;
\cref{fig:supp_matched}b). None of these sentences appears in any extraction prompt.

\begin{figure}[t]
\centering
\includegraphics[width=0.862\linewidth]{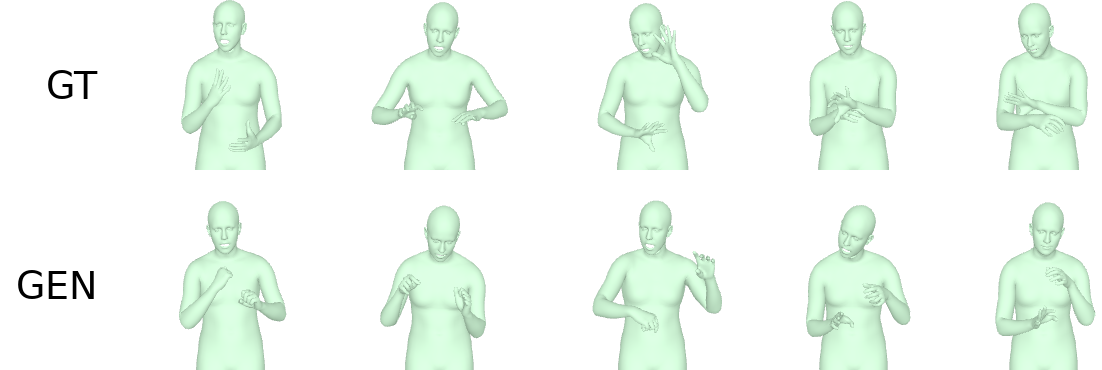}\\
{\footnotesize (a) Phoenix-2014T, typical non-formulaic sentence: $0.98$}\\[3pt]
\includegraphics[width=0.869\linewidth]{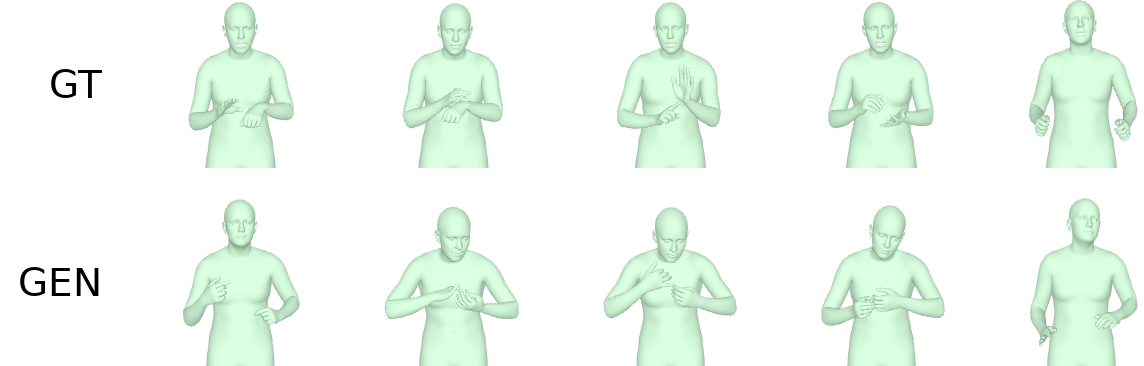}\\
{\footnotesize (b) CSL-Daily, at the generation median: $0.74$}
\caption{Additional DTW-matched frame pairs (GT top, generation
bottom), complementing Fig.~3. (a) articulated,
sentence-dependent signing whose trajectory does not follow this GT
rendition; (b) an unseen CSL-Daily sentence at the generation median.}
\label{fig:supp_matched}
\end{figure}

\section{Code-level audit of Progressive Transformers}
\label{sec:supp_pt}

A prominent Phoenix SLP system, Progressive
Transformers~\cite{saunders2020progressive}, reports better BLEU
than our checkpoints. Reading the released code, we find three
inference/evaluation paths that leak ground-truth timing or length
into the reported protocol; the two most important are (L1) the
predicted counter scalar is overwritten by the GT counter at every
decoder step, and (L2) the inference output length is set to the GT
sequence length rather than learned via a stop signal.

\paragraph{PT on the clean Phoenix-2014T SMPL-X data (Sec.~6.5).}
We retrained the official PT code on the same SOKE SMPL-X Phoenix data
as our MDM and regression models (FK-lifted to the $44$-joint $3$D
space the MoAE consumes) and ran inference twice on the same
checkpoint. Under fair inference (counter and length governed by the
model's own predictions) PT collapses: $\rfaith = 0.9999$,
$\rdiv = 0.025$, $d_{\rm inter} = 0.08$ against a GT spread of $3.26$.
Under the paper's leaky protocol (GT counter and length supplied at
inference) $\rfaith = 0.9974$ and $\rdiv = 0.223$: the leak inflates
apparent output diversity $\sim$$9\times$ while leaving faithfulness at
chance. MDM ($0.839$) and the clean regression model ($0.811$) are
faithful on the identical data. The remainder of this section reports
the earlier leaky-vs-fair experiment on PT's own pose representation,
which reaches the same conclusion.

\paragraph{Leaky-vs-fair experiment.}
We retrained PT's released code on Phoenix-2014T at the paper's native
data scale and ``best T2P'' configuration, with no changes to
architecture, loss, training, or inference logic, then ran inference
twice on the \emph{same} checkpoint over all $632$ test sentences:
once with leaks (L1)+(L2) intact (``leaky''), once with all three
leaks off (``fair'', counter and length governed purely by the
model's predictions). Outputs were encoded with the Phoenix MediaPipe
MoAE (\cref{sec:supp_impl}) and $d_{\rm inter}$/$\rdiv$/$\rfaith$
computed on the full test split (\cref{tab:supp_pt_audit}).

\paragraph{Representation and extraction.}
The independent PT retrainings cited in the main text use
OpenPose-derived 3D joints~\cite{huang2021fast, xie2024g2pddm} or a
SMPL-X representation~\cite{yin2024t2sgpt}, and PT's original
pipeline uses OpenPose. For this experiment we extract Phoenix poses
with MediaPipe Holistic, because OpenPose's released codebase has very
limited maintenance and is difficult to deploy in a modern
environment. This extraction is imperfect: across the Phoenix train
split, $28.6\%$ of left-hand and $23.9\%$ of right-hand frames are
returned with no detection and are interpolated (see hand-detection
notes below). We therefore use the leaky-vs-fair result only as a
\emph{within-checkpoint} comparison (the checkpoint, and any
extraction artefact, is held fixed across the two inference passes,
so the contrast still isolates the leak), and do not report absolute
PT performance from this retraining; the SMPL-X retraining above
uses the same clean data as every other Phoenix model in the paper.

\begin{table}[t]
\centering
\small
\setlength{\tabcolsep}{4pt}
\caption{Toggling PT's inference-time GT-timing/length leak on a
\emph{single} retrained checkpoint (MediaPipe representation).
$d_{\rm inter}$ is the mean pairwise L2 distance between generated
samples in the MoAE latent; $\rdiv$ normalises it by the GT spread.
The checkpoint is held fixed, so the comparison isolates the leak's
effect.}
\label{tab:supp_pt_audit}
\begin{tabular}{lcc}
\toprule
Metric & Fair (no leak) & Leaky (paper) \\
\midrule
$d_{\rm inter}$            & $0.031$ & $0.207$ \\
$d_{\rm inter\text{-}GT}$  & $2.321$ & $2.321$ \\
$\rdiv$ (inter / GT)       & \textbf{0.013} & \textbf{0.089} \\
$\rfaith$                  & $0.999$ & $0.996$ \\
\bottomrule
\end{tabular}
\end{table}

\paragraph{Findings.}
(i) \textbf{The leak measurably changes PT's output.} Toggling the
GT-timing/length leak on the same trained checkpoint inflates the
inter-sample spread $d_{\rm inter}$ from $0.031$ to $0.207$ and the
diversity ratio $\rdiv$ from $0.013$ to $0.089$, a roughly
sevenfold change (\cref{tab:supp_pt_audit}), while $\rfaith$ stays at
chance ($0.999 \to 0.996$). Because the checkpoint is held fixed
across the two inference passes, this isolates the leak: the
GT-aligned timing it supplies materially affects the generated motion
and any statistic computed from it, without making the motion any
more faithful to its sentence.
(ii) \textbf{Fair inference does not learn end-of-sequence in our
text-to-pose setting.} Without the GT counter, the predicted counter
never crosses the stop threshold: \emph{every one} of the $632$ test
samples runs to the $300$-frame cap, while GT lengths range from $3$
to $89$ frames. Under text conditioning, PT's continuous-output
decoder relies on (L1) for inference-time length supervision; a
gloss-conditioned PT, whose input already encodes sign count, retains
a partial length signal (leak-free gloss-to-pose PT reaches BLEU-4
$4.04$; \citealp{huang2021fast}).

The rest of this section gives the full L1--L3 enumeration, the
training setup for the leaky-vs-fair experiment, and the supporting
pose-extractor, hand-detection, and bug-fix notes needed to make the
released code execute under modern tooling.

\paragraph{Three inference- and evaluation-time leaks.}
Inspection of the released repository identifies the following three
paths through which GT timing or length information enters PT's
reported numerical protocol:
\begin{enumerate}
\item[(L1)] \texttt{search.py}:55--63: at every decoder step the
  predicted counter scalar $\hat{c}_t$ is overwritten with the GT
  counter $c^*_t$ before being fed back. The author comment in the file
  reads: ``\emph{Drive the timing by giving the GT timing}''.
\item[(L2)] \texttt{search.py}:46: the inference output length is
  assigned from the GT target length rather than learned via a stop
  signal.
\item[(L3)] \texttt{Configs/Base.yaml}:25,29: the only
  validation/early-stopping metric is the DTW score itself, so the
  reported quantitative results come from this GT-aligned distance.
\end{enumerate}

\paragraph{Training setup and toggle list.}
Training matched the paper's data scale ($\sim$$7\,000$ training
sequences) and the published ``best T2P'' configuration ($2$
encoder/decoder layers, $4$ heads, $512$ embedding dim, MSE loss,
future-prediction $\mathrm{FP}{=}10$, Gaussian-noise augmentation rate
$5$). Training reached $\mathrm{lr}_{\min}$ at step $1450$, with best
validation DTW $= 15.76$ at step $450$. The ``leaky'' inference
configuration in \cref{tab:supp_pt_audit} leaves (L1)+(L2) intact
(matching the paper's reported protocol up to BLEU back-translation);
the ``fair'' configuration disables all three leaks listed above
(counter is purely autoregressive, output length is governed by a
predicted-counter stop rule capped at the global $300$-frame limit,
and the DTW-aligned distance is no longer used as the
model-selection metric). Both inference passes were run over all
$632$ test sentences on the same trained checkpoint.

\paragraph{Pose extraction divergence from the original paper.}
Saunders \emph{et al.}~\cite{saunders2020progressive} use OpenPose
followed by 2D-to-3D inverse-kinematics lifting; we substitute
MediaPipe Holistic upper-body keypoints because the OpenPose pipeline
is impractical to re-deploy on contemporary GPU toolchains and the
extraction is not released. The two extractors differ in keypoint set
and 2D-vs-3D representation, but the total paired-sample budget is
identical at $\sim$$7\,000$ training sequences, so the
leak audit still applies; what we cannot precisely
reproduce are the absolute pose-space FID values (which depend on
extraction).

\paragraph{Hand-detection failure rate.}
MediaPipe Holistic produces all-zero hand keypoints when detection
confidence is below threshold. Across the Phoenix train split, the
mean per-video fraction of fully-missing left-hand frames is $28.6\%$
(max $100\%$); for the right hand $23.9\%$ (max $84\%$). We
linearly-interpolate within each clip and fall back to the corresponding
body wrist when an entire hand is missing throughout, so PT does not
train against zero-hand targets.

\paragraph{Bug fixes required to run the released code under a
modern deep-learning toolchain.} We applied six small patches to make
the released code execute under the current toolchain. \textbf{None of
the patches touch the architecture, loss, training loop, validation
loss/DTW computation, model selection, or inference/search code}: the
model and training-relevant files (\texttt{model.py},
\texttt{transformer\_layers.py}, \texttt{encoders.py},
\texttt{decoders.py}, \texttt{embeddings.py}, \texttt{loss.py},
\texttt{search.py}, \texttt{data.py}, \texttt{batch.py},
\texttt{vocabulary.py}) are byte-for-byte the released versions.
The patches are: removing the deprecated \texttt{verbose} kwarg from
\texttt{ReduceLROnPlateau} (\texttt{builders.py:137}); fixing an
undefined \texttt{train\_output} reference (\texttt{prediction.py:73});
removing a redundant FP-cut on the inference output already cut in
\texttt{search.py:greedy} (\texttt{prediction.py:74}); making
\texttt{plot\_videos.py:47} adaptive to MediaPipe's $67$-joint layout;
wrapping \texttt{produce\_validation\_video} in try/except to swallow
bone-graph mismatches at validation-video render time (the
\texttt{\_save\_checkpoint} call lies outside the try block, so model
selection is unaffected); and adding \texttt{weights\_only=False} to
\texttt{torch.load} (\texttt{helpers.py:202}). Total diff vs the
upstream repo: $32$ lines added, $20$ lines removed across $5$ files.

\paragraph{Reproducibility.}
The patched code, training and audit scripts, configurations, and all
per-sample inference outputs are available alongside the paper.
Reproducing \cref{tab:supp_pt_audit} end-to-end requires only a single
GPU.

\fi

\end{document}